\documentclass[sigconf]{acmart}

\AtBeginDocument{%
  }

\setcopyright{acmlicensed}
\copyrightyear{2018}
\acmYear{2018}
\acmDOI{XXXXXXX.XXXXXXX}
\acmConference[Conference acronym 'XX]{Make sure to enter the correct
  conference title from your rights confirmation emai}{June 03--05,
  2018}{Woodstock, NY}
\acmISBN{978-1-4503-XXXX-X/18/06}

\usepackage{amsmath,amsfonts,bm}

\def\eqref#1{equation~\ref{#1}}

\def\1{\bm{1}}

\DeclareMathAlphabet{\mathsfit}{\encodingdefault}{\sfdefault}{m}{sl}
\SetMathAlphabet{\mathsfit}{bold}{\encodingdefault}{\sfdefault}{bx}{n}

\newcommand{\E}{\mathbb{E}}

\newcommand{\R}{\mathbb{R}}

\usepackage{xspace}
\usepackage{booktabs}
\usepackage{graphicx}
\usepackage{multicol, multirow, xcolor, threeparttable}
\usepackage{gensymb}
\usepackage{hyperref}
\usepackage{url}
\usepackage{makecell}
\newcommand{\emoji}[2][1.2em]{\raisebox{-0.2\height}{\includegraphics[height=#1]{#2}}}
\newcommand{\NAME}{Baguan\xspace}

\def\0{\mathbf{0}}

\def\R{\mathbb{R}}
\def\E{\mathbb{E}}
\def\N{\mathcal{N}}

\def \X {\mathbf{X}}

\def \Z {\mathbf{Z}}
\begin{document}


\title{Utilizing Strategic Pre-training to Reduce Overfitting: Baguan - A Pre-trained Weather Forecasting Model
}

\author{Peisong Niu}
\authornote{Authors contributed equally to this research.}
\email{niupeisong.nps@alibaba-inc.com}
\affiliation{%
  \institution{DAMO Academy, Alibaba Group}
  \city{Hangzhou}
  \country{China}
}

\author{Ziqing Ma}
\authornotemark[1]
\email{maziqing.mzq@alibaba-inc.com}
\affiliation{%
  \institution{DAMO Academy, Alibaba Group}
  \city{Hangzhou}
  \country{China}
}

\author{Tian Zhou}
\authornotemark[1]
\email{tian.zt@alibaba-inc.com}
\affiliation{%
  \institution{DAMO Academy, Alibaba Group}
  \city{Hangzhou}
  \country{China}
  \postcode{}
}

\author{Weiqi Chen}
\email{jarvus.cwq@alibaba-inc.com}
\affiliation{%
  \institution{DAMO Academy, Alibaba Group}
  \city{Hangzhou}
  \country{China}
  \postcode{}
}

\author{Lefei Shen}
\email{shenlefei.slf@alibaba-inc.com}
\affiliation{%
  \institution{DAMO Academy, Alibaba Group}
  \city{Hangzhou}
  \country{China}
  \postcode{}
}

\author{Rong Jin}
\email{rongjinemail@gmail.com}
\affiliation{%
  \institution{DAMO Academy, Alibaba Group}
  \city{Bellevue}
  \country{USA}
}

\author{Liang Sun}
\email{liang.sun@alibaba-inc.com}
\authornote{Corresponding authors.}
\affiliation{%
  \institution{DAMO Academy, Alibaba Group}
  \city{Bellevue}
  \country{USA}
}

\begin{abstract}

Weather forecasting has long posed a significant challenge for humanity. While recent AI-based models have surpassed traditional numerical weather prediction (NWP) methods in global forecasting tasks, overfitting remains a critical issue due to the limited availability of real-world weather data spanning only a few decades. Unlike fields like computer vision or natural language processing, where data abundance can mitigate overfitting, weather forecasting demands innovative strategies to address this challenge with existing data. In this paper, we explore pre-training methods for weather forecasting, \textbf{finding that selecting an appropriately challenging pre-training task introduces locality bias, effectively mitigating overfitting and enhancing performance}. We introduce \NAME, a novel data-driven model for medium-range weather forecasting, built on a Siamese Autoencoder pre-trained in a self-supervised manner and fine-tuned for different lead times. Experimental results show that \NAME outperforms traditional methods, delivering more accurate forecasts. Additionally, the pre-trained \NAME demonstrates robust overfitting control and excels in downstream tasks, such as subseasonal-to-seasonal (S2S) modeling and regional forecasting, after fine-tuning.




\end{abstract}





\begin{teaserfigure}
    \centering
    \includegraphics[width=0.8\textwidth]{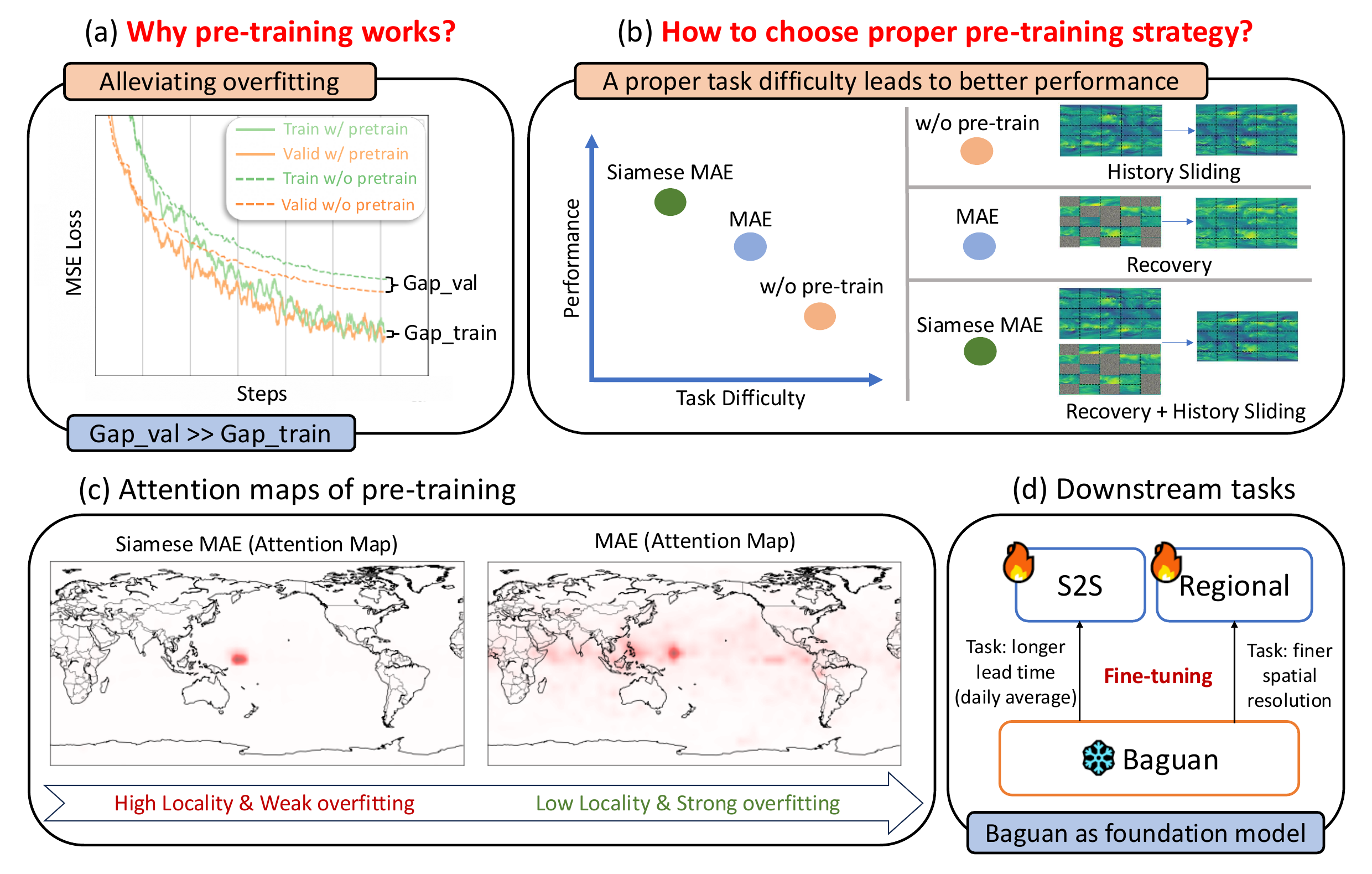}
    \vskip -0.1in
    \caption{A brief review. 
    (a) Pre-training can effectively alleviate overfitting.
    (b) A proper task difficulty leads to better performance.
    (c) Attention maps of Siamese MAE and MAE. (d) Downstream tasks for \NAME through fine-tuning.} 
    \label{fig:intro}
\end{teaserfigure}

\maketitle



\section{Introduction}

Data-driven weather forecasting models have significantly advanced the boundaries of meteorological science~\cite{graphcast,pangu_nature,Fuxi_nature,Chen2023FengWuPT}. 
These advanced 
models are typically trained on the ERA5 
dataset
~\cite{hersbach2020era5}, produced by the European Centre for Medium-Range Weather Forecasts (ECMWF). Notably, these approaches have consistently outperformed traditional numerical weather prediction (NWP) techniques in both accuracy in standard forecast metrics and inference speeds for operational use~\cite{rasp2024weatherbench2benchmarkgeneration, FoundationWeatherASurvey}. 
However, as data-driven models grow in scale, the risk of overfitting becomes a significant concern. Unlike fields such as Computer Vision (CV) and Natural Language Processing (NLP) which benefit from the availability of large scale datasets, weather forecasting is constrained by limited real-world data, exemplified by the ERA5 dataset, which spans only a few decades~\cite{hersbach2020era5}. 

\begin{table*}[h]
    \centering
    \caption{Summary of SOTA AI-powered Global Weather Forecasting Methods and the Use of Pre-training.}
    \resizebox{\textwidth}{!}{
        \begin{tabular}{|c|c|c|c|c|c|c|c|c|c|c|}
            \hline
            \textbf{Method} & Pangu\cite{pangu_nature} & FuXi\cite{Fuxi_nature} & GraphCast~\cite{graphcast} & FengWu~\cite{Chen2023FengWuPT} & FCN~\cite{pathak2022fourcastnet} &Aurora~\cite{Bodnar2024AuroraAF}&Climax~\cite{Nguyen2023ClimaXAF}&Prithvi~\cite{schmude2024prithvi} & \textcolor{red}{Ours: MAE}~\cite{MAEareScalableVisionLearners,man2023w} &\textcolor{red}{Ours: Siamese MAE}\\
            \hline
            \textbf{Pre-training Used} & \emoji{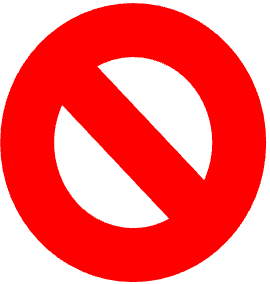}  & \emoji{figures/cha.png}  & \emoji{figures/cha.png}  & \emoji{figures/cha.png}  & \emoji{figures/cha.png}  &\emoji{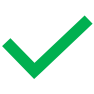}& \emoji{figures/gou.png} & \emoji{figures/gou.png} &\emoji{figures/gou.png}&\emoji{figures/gou.png} \\
            \hline
            \textbf{Impact of Pre-training} & - & - & - & - & - &Positive& Positive &Negative & Mildly Positive &\textcolor{red}{Positive} \\
            \hline
            \textbf{Free of Additional Data}& \emoji{figures/gou.png} & \emoji{figures/gou.png} &\emoji{figures/gou.png} & \emoji{figures/gou.png} & \emoji{figures/gou.png} &\emoji{figures/cha.png} & \emoji{figures/cha.png} & \emoji{figures/gou.png} & \emoji{figures/gou.png}&\emoji{figures/gou.png} \\
            \hline
        \end{tabular}
    }
    \label{tab:methods}
\end{table*}

In weather forecasting, overfitting is often linked to inaccurate interactions. The vast input size, which encompasses the entire Earth, can cause the model to learn arbitrary relationships, especially when global attention mechanisms are employed. Beyond augmenting training data, such as introducing extensive climate simulation datasets~\cite{Bodnar2024AuroraAF}, researchers have pursued two main strategies to address this issue. \textbf{The first strategy involves designing models that incorporate locality bias}, such as those based on CNNs~\cite{cheon2024karina}, GNNs~\cite{graphcast,lang2024aifsecmwfsdatadriven}, or local attention mechanisms~\cite{fuxi2.0, Chen2023FengWuPT, pangu_nature}. These models have been thoroughly explored in the context of weather forecasting. \textbf{The second strategy focuses on using pre-training tasks to introduce locality bias into 
models}, which generally offers more adaptability and flexibility than the model design approach. However, it is surprising that the effectiveness of this method has been largely underreported in previous research, with the exception of a few studies~\cite{man2023w, Nguyen2023ClimaXAF}. Notably, most widely used weather forecasting models only utilize the prediction task as their primary focus, essentially treating it as a pre-training task~\cite{Nguyen2023ClimaXAF, Bodnar2024AuroraAF}. They do not employ an independent pre-training stage to enhance the subsequent forecasting task. We believe the community might have attempted this approach and encountered challenges or mixed results. As highlighted in \cite{schmude2024prithvi}, the MAE pre-training stage significantly hinders subsequent weather forecasting performance. In contrast, \cite{man2023w} documents a positive impact of MAE. Our experiments reveal a subtle yet positive enhancement in downstream weather forecasting tasks, though this improvement might be easily overlooked when utilizing a simplistic MAE pre-training method. This may explain why few weather forecasting models utilize an independent pre-training stage as shown in Table~\ref{tab:methods}. Given that the pre-training stage is model-agnostic, we believe this approach holds significant potential for controlling overfitting without altering the model design. It is also highly applicable to downstream tasks due to its simplicity and flexibity.

In this paper, we explore pre-training as a means to mitigate overfitting in weather forecasting models as shown in Fig.~\ref{fig:intro}(a). We focus on Masked Image Modeling (MIM)~\cite{MAEareScalableVisionLearners, xie2021simmim, gupta2023siamese} for weather data, recognizing the complexity of weather signals compared to images or videos, which makes MAE less effective. To tackle this, we use a Siamese MAE framework that introduces additional signals to gradually adjust pre-training difficulty (formally defined in Eq. \ref{eq:task_difficulty}), employing a dual objective of recovery and prediction tasks, as illustrated in Fig.~\ref{fig:intro}(b). Our approach, detailed through multi-stage tuning and comparative analyses with models like ClimaX~\cite{Nguyen2023ClimaXAF}, shows that pre-training significantly enhances overfitting control, particularly in climate-related tasks with limited data. 
And the effective control of overfitting is achieved by introducing a relatively strong local inductive bias through a model-agnostic design, as illustrated in Fig. \ref{fig:intro}(c).

Building upon the analysis of pre-training, 
we introduce \NAME, a Transformer-based model that leverages a pre-training $\&$ fine-tuning paradigm with Siamese MAE. Due to the comprehensive training on vast datasets, \NAME emerges as an ideal foundational model for diverse downstream tasks. Beyond the global weather forecasting task, two prominent applications that leverage \NAME as their base model include sub-seasonal to seasonal (S2S) climate forecasting \cite{subseasonal2seasonalprediction, guo2024maximizing} and regional fine-scale forecasting. Our contributions can be summarized as follows:

\begin{enumerate}
    \item We conduct an in-depth study on pre-training methods to evaluate their impacts on global weather forecasting, particularly controlling overfitting and analyzing the link between task difficulty and performance. The results demonstrate that our Siamese MAE surpasses the standard MAE and forecasting-based pre-training methods through effectively tuning task difficulty.
    \item We present \NAME, which excels in medium-range global forecasting tasks and outperforms both Pangu-Weather and IFS by utilizing a three-stage tuning process and a rolling-based forecasting method. 
    \item \NAME showcases its capability as a foundational model for a range of downstream tasks. It achieves SOTA performance in S2S forecasting, surpassing ECMWF-S2S, while also delivering robust results in high-resolution regional forecasting within real-world, industry-level weather application systems.

\end{enumerate}

\section{Related Work}

Deep global models leverage a data-driven approach to learn the medium-range spatiotemporal relationships among atmospheric variables. Within this context, two primary frameworks have gained significant attention: the Graph Neural Networks (GNN) family and the Transformers family. The GNN family, which includes models such as  GraphCast~\cite{graphcast} and AIFS~\cite{lang2024aifsecmwfsdatadriven}, represents the Earth's surface as an irregular and uniformly spaced grid instead of a regular rectangular grid. This innovative representation enables more flexible and efficient modeling of the Earth's complex geometry. In the Transformer family, ClimaX~\cite{Nguyen2023ClimaXAF} and Stormer~\cite{stomer} originate from the standard Vision Transformer (ViT). ClimaX, in particular, introduces a pre-training framework on climate simulation data. Additionally, models such as Pangu-Weather~\cite{pangu_nature}, FuXi~\cite{Fuxi_nature}, and FengWu~\cite{Chen2023FengWuPT} leverage the advantages of the Swin Transformer, which effectively captures local features and optimizes memory usage.

Since most deep weather forecasting models leverage ViT, the latest advancements in CV can provide valuable inspiration and insights. Within the field of CV, masked image modeling has seen rapid development. MAE~\cite{MAEareScalableVisionLearners} and SimMIM~\cite {xie2021simmim} are two notable approaches that involve masking or corrupting a specific portion of information within an image. The model is then tasked with recovering this masked portion, thereby promoting the learning of valuable image representations. Siamese MAE~\cite{gupta2023siamese} introduces the MAE method for video modeling by incorporating two frames and masking a substantial portion of information in the second frame. Additionally, \cite{amos2024never} has demonstrated that the integration of data-driven priors through pre-training is crucial for achieving reliable performance estimation in a cost-effective manner.


\section{Methodology}
\begin{figure*}[t]
    \centering
    \includegraphics[width=0.85\textwidth]{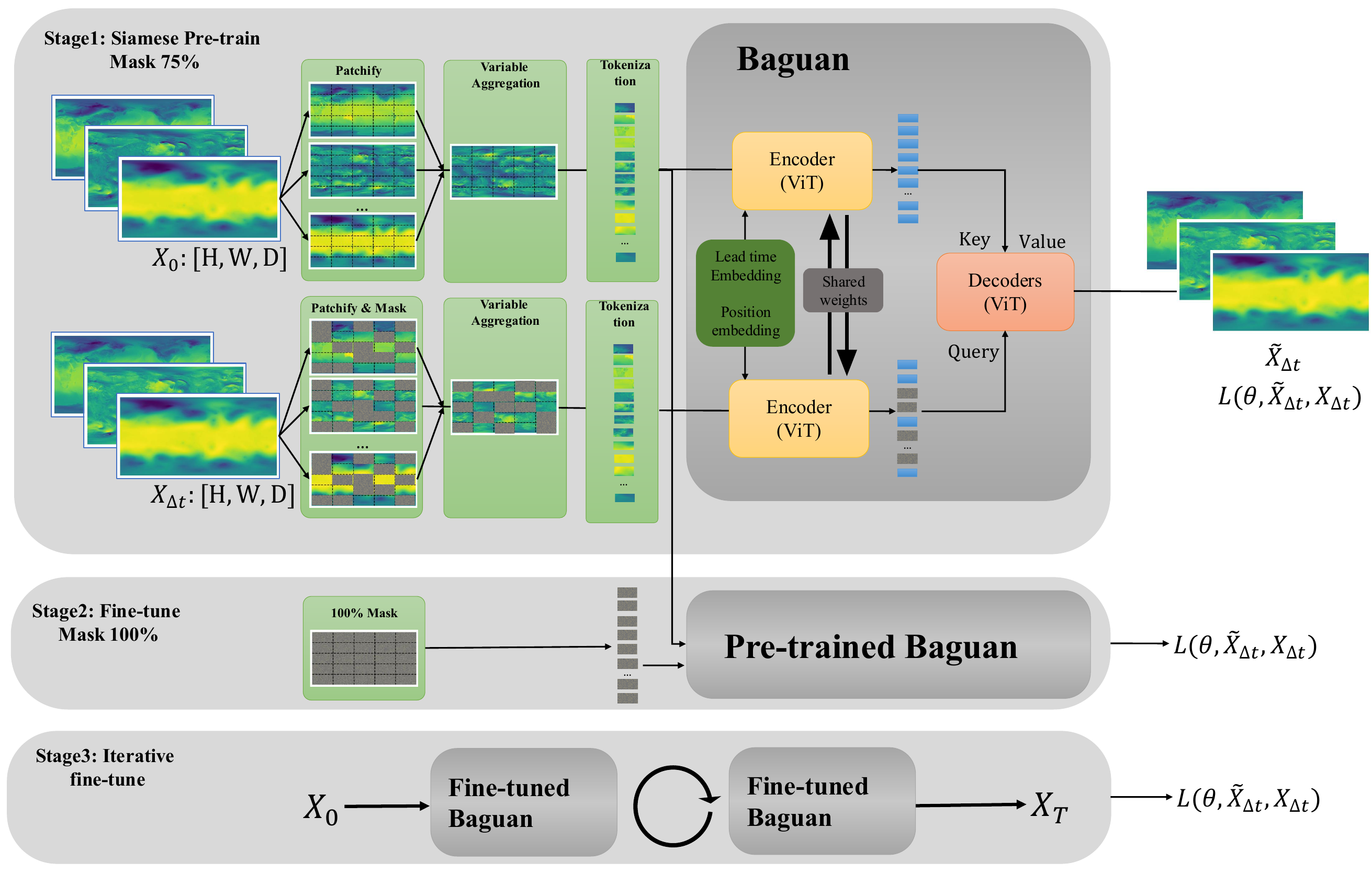}
    \caption{Overview of \NAME's architecture and its three training stages. The process begins with the aggregation of the initial state $\X^{t_0}$ and a masked state $\X^{t_0+\Delta t}_{masked}$ 
    into a single channel respectively, which are then input into a weight-sharing encoder. The two representations are subsequently combined through a cross-self decoder and a prediction head to produce the reconstruction or forecasting results $\X^{t0+\Delta t}$.}
    \label{fig:model_arch}
\end{figure*}

\subsection{Problem Formulation}
\NAME is an advanced data-driven framework designed to generate accurate global medium-range weather forecasts. \NAME takes as input the atmospheric state at time $t_0$, represented as $\X^{t_0} \in \R^{V \times H \times W}$. $V$ denotes the number of atmospheric variables considered in the input, while $H$ and $W$ represent height and width, respectively. In practical scenarios with a spatial resolution of $0.25^\circ$, $H=721$ and $W=1440$ for the Earth. \NAME aims to establish the mapping: 
\begin{equation}
    \tilde{\X}^{t_0+\Delta t} = \mathrm{\NAME}(\X^{t_0}),
\end{equation} 
where $\Delta t$ denotes time interval (6 hours). In fact, \NAME produces 6-hourly forecasts in an autoregressive manner: $\tilde{\X}^{t_0+n\Delta t} = \mathrm{\NAME}^{(n)}(\X^{t_0})$, where $n$ indicates the number of rolling steps. In this way, it can make prediction extending up to a maximum of 7 days.

\subsection{The Architecture of \NAME}

Fig.~\ref{fig:model_arch} provides an overview architecture of \NAME. 
\NAME is a Siamese MAE pre-trained through a self-supervised learning paradigm. Its architecture consists of a variable aggregation $\&$ tokenization module, a ViT-like encoder, and a cross-self decoder. A details model configure can be found in Appendix Table~\ref{tab:model_details}.



\subsubsection{Variable Aggregation \& Tokenization}
Unlike images, which typically consist of three channels, atmospheric states often encompass more than seventy variables, which presents significant challenges. Therefore, we employ a weather-specific embedding~\cite{Nguyen2023ClimaXAF, stomer} instead of the standard patch embedding used in ViT. The specific weather embedding module is composed of a variable aggregation module and a tokenization module. Their objective is to transform the image-like input $\X: V \times H \times W$ into a sequence of tokens.

In the variable aggregation module, the input tensor $\X: V \times H \times W$ is processed through a series of variable-specific embedding layers, resulting in a tensor of shape $V \times H \times W \times D$. Subsequently, a cross-attention mechanism is employed, utilizing a learnable query with dimensions $1 \times H \times W \times D$ to aggregate the embedded tensor into a tensor of shape $1 \times H \times W \times D$.

In the variable tokenization module, a patchify method is applied to convert this aggregated tensor into a sequence of tokens with shape $(H/p * W/p) \times (D * p^2)$, where $p$ is patch size.

\subsubsection{Encoder}
In the encoder, the tokens generated by the initial state $\X_1$ at time $t_0$ and the target state $\X_2$ at time $t_0 + \Delta t$ are subsequently processed by a stack of Transformer blocks, producing output tokens denoted as $\Z_1$ and $\Z_2$. The encoder employs the standard ViT architecture. During the pre-training stage, the initial state $\X_1$ remains unmasked, whereas the target state $\X_2$ is partially masked with Gaussian noise with a ratio of 75\%. In contrast, during the fine-tuning and inference stages, the target state $\X_2$ is replaced by pure Gaussian noise. Additionally, lead time embedding and positional embedding are incorporated into the tokens to provide contextual information. Notably, the two input frames share the same encoder weights, enabling the model to learn shared representations across both inputs.

\subsubsection{Decoder}
The decoders used in the pre-training (stage 1) and fine-tuning (stages 2 and 3) processes exhibit slight differences, and the weight of the decoder in pre-training will not be carried over to the fine-tuning stages.
The purpose of pre-training is to slowly reconstruct $\X^{t_0+\Delta t}$ with limited cues, while fine-tuning is aimed at forecasting $\X^{t_0+\Delta t}$ without any cues, based solely on the initial state $\X^{t_0}$.
Therefore, we employ two distinct decoders during the pre-training and fine-tuning stages.

\subsubsection{Cross-self Decoder} 
A cross-self decoder architecture is employed, which consists of a stack of blocks. Each block is composed of a cross-attention layer and a self-attention layer. The entire process can be represented by:
\begin{equation}
    \label{eq:cross_attention}
    Z_2 = Z_2 + \mathrm{MultiHeadAttention}(Z_2 W^Q_1, Z_1 W^K_1, Z_1 W^V_1),
\end{equation}
\begin{equation}
    \label{eq:self-attention}
    Z_2 = Z_2 + \mathrm{MultiHeadAttention}(Z_2 W^Q_2, Z_2 W^K_2, Z_2 W^V_2)
\end{equation}
\begin{equation}
    \label{eq:ffn}
    Z_2 = Z_2 + \mathrm{FFN}(Z_2),
\end{equation}
where $W^Q_1, W^K_1, W^V_1, W^Q_2, W^K_2, W^V_2$ are the parameter matrices of cross-attention layer and self-attention layer. $\mathrm{MultiHeadAttention}(\cdot)$ and $\mathrm{FFN}(\cdot)$ are the functions of multi-head attention module and feed-forward networks.


\subsubsection{Adaptive Layer Normalization}
Adaptive layer normalization (AdaLN)~\cite{Peebles2022ScalableDM, perez2018film} is a widely used technique
and has been successfully integrated into weather forecasting~\cite{stomer}.
In this paper, we employ AdaLN, a mechanism where the parameters of the normalization layer are learned from the embedding of $\Delta t$ in the Transformer blocks rather than traditional additive lead time embedding.

\subsection{Three Training Stages of \NAME}

As shown in Fig. \ref{fig:model_arch}, \NAME undergoes three training stages. Initially, the Siamese MAE is pre-trained on the Siamese masked image modeling (SMIM) tasks. Following this, \NAME is fine-tuned for specific lead times of 1 hour, 6 hours or 24 hours. The final stage involves iterative fine-tuning using multi-step data. 

\subsubsection{Stage 1: Pre-training with Siamese MAE}

Siamese network is a weight-sharing neural network that serves as a crucial element in modern contrastive representation learning methods~\cite{Chen2020ExploringSS}. 
Specifically, we employ the asymmetric masking strategy where no patches (0\%) in $\X^{t_0}$ are masked, and a high ratio ($r$=75\%) of patches in $\X^{t_0+\Delta t}$ are masked. The task difficulty can be adjusted by varying the masking ratio $r$. Notably, the task difficulty is inversely proportional to the number of tokens provided. For two input frames corresponding to time steps $t$ and $t-1$, the task difficulty ($D_{\rm{task}}$) during pretraining is described as:
\begin{equation}
    D_{\rm{task}} \propto \frac{1}{\lambda_t (1 - r_t) + \lambda_{t-1} (1 - r_{t-1})}
\label{eq:task_difficulty}
\end{equation}
where $\lambda_t > 0$ and $\lambda_{t-1} > 0$ denote the information contribution of frames $t$ and $t-1$, respectively, in reconstructing the unmasked tokens. $r_t$ and $r_{t-1}$ represents the masking ratios for frames $t$ and $t-1$, respectively. In the case of MAE, we have $r_{t-1} = 1$, whereas for Siamese MAE, $r_{t-1} = 0$.

Both $\X^{t_0}$ and $\X^{t_0+\Delta t}_{masked}$ share the encoder
weights and the representations are fed into the decoder to reconstruct $\X^{t_0+\Delta t}$. This task simplifies the forecasting process and encourages the model to concentrate on the changes occurring between two input frames.

\subsubsection{Stage 2: Fine-tuning on Fixed Lead Time}
After being pre-trained on the Siamese masked image modeling task, the \NAME encoder is subsequently fine-tuned using fixed lead times of 1 hour, 6 hours, or 24 hours.
In Stage 2, information about $\X^{t_0+\Delta t}$ is entirely unavailable, resulting in the input for frame 2 being effectively masked at a ratio of 100\%. Consequently, the frame 2 relies solely on positional and lead time embeddings.

\subsubsection{Stage 3: Iterative Fine-tuning}
To enhance \NAME's multi-step forecasting accuracy and reduce the error accumulation associated with simply rolling the model, we introduce an iterative fine-tuning strategy in Stage 3. This approach leverages an auto-regressive training scheme, where the model's output at each step is recursively fed back as input for subsequent predictions. Formally, for a given initial state $\X^{t_0}$, the prediction at time $t_0+n\Delta t$ is computed as:
\begin{align}
    \tilde{\X}^{t_0+n\Delta t} &= \mathrm{\NAME}({\ldots \mathrm{\NAME}(\X^{t_0})}) \\
    &= \mathrm{\NAME}^{(n)}(\tilde{\X}^{t_0}). \notag
\end{align}
During the inference stage, the auto-regressive strategy is similarly applied.


\subsection{Ensemble Strategy}
To deal with uncertainty in initial conditions and error accumulation introduced by longer forecast lead times, ensemble forecasting is commonly used in both traditional numerical weather prediction~\cite{coiffier2011fundamentals} and recently proposed AI-based methods~\cite{Fuxi_nature, pangu_nature, pathak2022fourcastnet}. 
Usually, it incorporates noise into the initial conditions to generate multiple predictions. Recently, Stormer~\cite{stomer} proposes a randomized iterative ensemble strategy that combines multiple models, including 6-hour and 24-hour models, to enhance long-term forecasting accuracy. However, its Best $m$ in $n$ strategy~\cite{stomer} still requires a comprehensive search through all $n$ combinations. To enhance the pruning of the search space, we propose a novel strategy outlined by the following rules:
\begin{enumerate}
    \item \textbf{Combination Rule}: We employ various combinations for lead times, such as [6, 6, 6, 6] and [24] for a 24-hour forecasting. The combinations are not restricted to identical lead times. For exmaple, 48 hours can be assembled in various ways, such as using four 6-hour and one 24-hour.
    \item \textbf{Pruning Rule}: The extended lead time can lead to an exponential increase in possible combinations. To address this, we implement a filtering mechanism that limits the number of iterations to enhance efficiency and accelerate the inference process.
\end{enumerate}
With this ensemble strategy, \NAME achieves an average improvement of approximately 2\% in longer lead-time forecasts, helping to reduce accumulated errors. Appendix \ref{app:ensemble} presents the detailed results of the ablation experiments.

\subsection{Pre-training as Regularization}
\label{sec:proof}

In this section, we develop a theoretical framework to explore how pre-training improves downstream task performance. Given the complexity of analyzing large deep learning models, we concentrate on linear regression to demonstrate how pre-training serves as a regularization mechanism, mitigating overfitting. We demonstrate that pre-training effects analogous to pruning the subspace spanned by eigenvectors with small eigenvalues, hence regularizing the final solution to rely primarily on the leading eigenvectors of the covariance matrix. This regularization enhances generalization when the target solution resides within the subspace of leading eigenvectors.

We begin with the standard setup of linear regression. For a set of training examples $\{(x_i, y_i)\}$ where $x_i \in \mathbb{R}^d$ and $y_i \in \mathbb{R}$, and a covariance matrix $\Sigma = \mathbb{E}[x_ix_i^{\top}]$, we consider an optimal weight vector $w_*$ in the subspace spanned by the top $K$ eigenvectors of $\Sigma$. Our analysis shows that pre-training with a denoising model using self-regression improves generalization performance.

We have the following key findings:

\begin{enumerate}
    \item Without pre-training, the error bound for the estimated weights, considering the top $K$ eigenvectors, is given by:
    \begin{equation}
        |w_1 - w_*| \leq \frac{\lambda}{R^2/2\sqrt{K} + \lambda} |w_*| + \frac{R}{n} + \sqrt{\frac{\sigma^2Rd^{1/2}}{n}\log\frac{1}{\delta}}. 
    \end{equation}
    \item By employing a pre-training method (denoising),  with high probability the error bound becomes:
    \begin{equation}
        |w_2 - w_*| \leq O\sqrt{\frac{1}{n}}.
    \end{equation}
\end{enumerate}
These findings indicate a significant improvement in the generalization error. Indeed, this analysis is performed on a simplified case, but it justifies our modeling based on pre-training for weather forecasting to some extend.
Detailed derivations and proofs supporting these results are provided in Appendix~\ref{appendix_provement}.

\section{Experiments}

\subsection{Experimental Settings}

\label{section:training_detail}
\subsubsection{Dataset}

We utilize the ERA5 dataset~\cite{hersbach2020era5} as the ground truth for model training and inferencing. Produced by the European Centre for Medium-Range Weather Forecasts (ECMWF), ERA5 is a global atmospheric reanalysis dataset providing detailed information on the Earth's climate and weather conditions from 1940 to the present. It includes a wide range of variables, such as temperature, humidity, precipitation, and mean sea level pressure, etc. This dataset is available at a high spatial resolution of $0.25^\circ$ latitude-longitude, roughly equivalent to approximately 31 kilometers, and encompasses 37 vertical pressure levels. As detailed in Appendix~\ref{app:datset}, we select a subset of 13 of these levels (50 hPa, 100 hPa, 150 hPa, 200 hPa, 250 hPa, 300 hPa, 400 hPa, 500 hPa, 600 hPa, 700 hPa, 850 hPa, 925 hPa, 1000 hPa), 
along with four surface variables (2m temperature, u-component and v-component of 10m wind speed, and mean sea level pressure) and several statistic variables. 
Following Pangu-Weather~\cite{pangu_nature}, we use the data from 1979 to 2015 for training, 2016 to 2017 for validation, and 2018 for testing. 
Same as Pangu-Weather, our training stages involve using 1-hourly data, rather than 6-hourly in FuXi~\cite{Chen2023FuXiAC} and FengWu~\cite{Chen2023FengWuPT}.
During testing, forecasts are initialized at 00 and 12 UTC, a standard benchmark time for comparing forecast scores across weather forecasting models~\cite{rasp2024weatherbench2benchmarkgeneration}.


\subsubsection{Implementation Details}

The size of the input data from ERA5 is $V \times 1440 \times 721$. 
During the tokenization stage, we employ a patch size of $8 \times 8$, which results in 16380 tokens. \NAME consists of encoders with a depth of 16, and a hidden dimension of 2048, as well as decoders with a depth of 4 and hidden dimension of 1024 both for pre-training and fine-tuning. A batch size of 1 is employed on each GPU. The model configuration and details of each stage can be found in Appendix~\ref{app:model_config} and Appendix~\ref{app:imp}. Furthermore, we intend to open-source the checkpoint along with all training and evaluation codes, believing it will serve as a valuable baseline for the community.

\subsubsection{Evaluation Metrics \& Training Objective}

Following prior work~\cite{pangu_nature, graphcast}, two specific quantitative metrics are employed to assess the accuracy of forecasting: the Latitude-Weighted Root Mean Square Error (RMSE) and the Latitude-Weighted Anomaly Correlation Coefficient (ACC). 

For optimization, we use the Mean Square Error (MSE) loss for pre-training and Mean Absolute Error (MAE) loss for fine-tuning. More details can be found in Appendix~\ref{app:metric}.



\subsection{The Effectiveness of Siamese MAE Pre-training}

\begin{figure}[h]
    \centering
    \includegraphics[width=0.49\columnwidth]{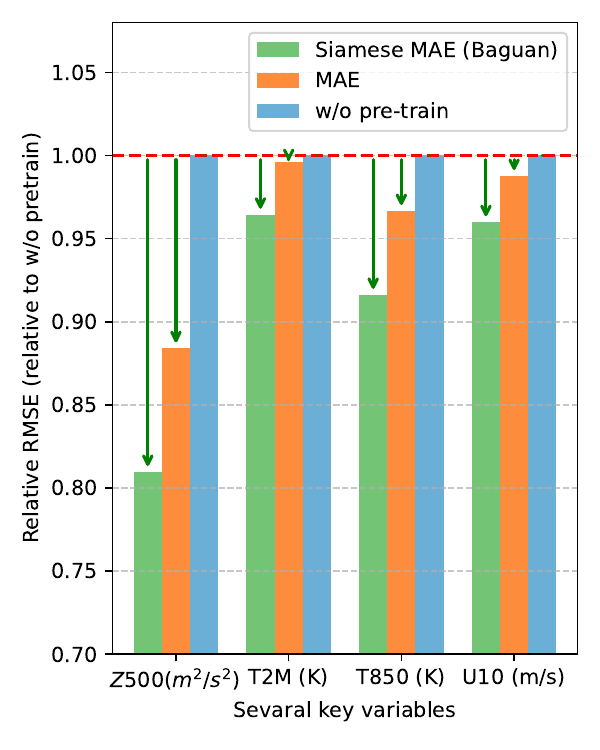}
    \includegraphics[width=0.49\columnwidth]{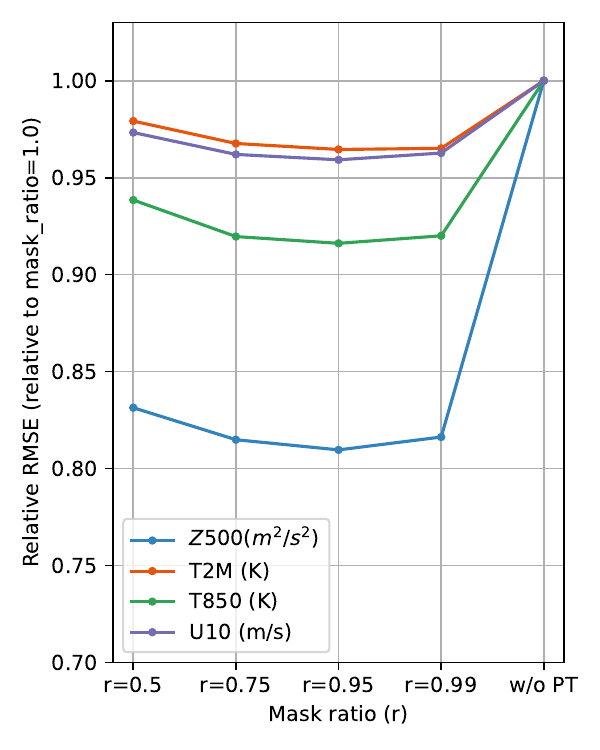}
    \caption{\small(Left): Comparison of relative RMSE for 6-hour forecasts across various weather variables, including Z500, T2M, T850, and U10. The RMSE values are presented relative to the baseline model without pre-train. Different pre-training methods are evaluated: \NAME (Siamese MAE) and MAE. (Right): Comparison of relative RMSE across different masking ratios ranging from 0.5 to 0.99. The RMSE values are presented relative to the version without pre-train (w/o PT).}
    \label{fig:ablation}
\end{figure}

\subsubsection{Empirical Comparison of Pre-training Approaches}

To evaluate the effectiveness of different pre-training approaches, we conducted a comparison between Siamese MAE~\cite{gupta2023siamese} and MAE~\cite{he2022masked}. The without pre-train version served as a baseline to normalize the results (RMSE) obtained from both Siamese MAE and MAE pre-training. Among these pre-training tasks, the w/o pre-train version presents the most challenging scenario, as it has no information regarding the second frame. In contrast, the MAE focuses on reconstructing the current frame while preserving 25\% of the patches. The Siamese MAE, on the other hand, represents the simplest task, as it utilizes information from both the current and 25\% of the second frames. 

Fig.~\ref{fig:ablation} (Left) provides the results using the $1.40625^\circ$ resolution data for 6-hour prediction. The results indicate that \NAME, when utilizing Siamese MAE pre-training, outperforms the w/o pre-training version by an average improvement of \textbf{10.1\%} in RMSE across four key variables. In contrast, \NAME with standard MAE pre-training achieves only a \textbf{4.5\%} average improvement over the w/o pre-training version, underscoring the superior effectiveness of the Siamese MAE approach.

The primary purpose of introducing pre-training is to mitigate overfitting, as discussed earlier in Fig.~\ref{fig:intro}(a). To quantify this effect, we calculate the mean MSE over the last 50 training steps. The pre-trained model exhibits a train-validation loss gap of $3.6\mathrm{e}{-4}$, compared to $4.0\mathrm{e}{-4}$ for the non-pre-trained model, underscoring the role of pre-training in reducing overfitting.


\subsubsection{Spectrum Analysis of Learned Representations}

In Sec \ref{sec:proof}, our theoretical analysis demonstrates that pre-training mitigate overfitting by regularizing the solution to rely on leading eigenvectors of the covariance matrix.
\begin{figure}[h]
    \centering
    \includegraphics[width=0.8\columnwidth]{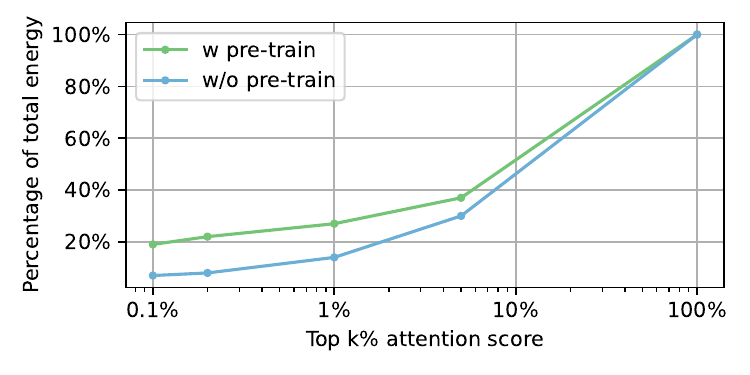}
    \caption{The total spectrum energy of top k\% attention scores. The total number of tokens is about 16,000. The sum of energy is 1.}
    \label{fig:spectral_analysis}
\end{figure}
For Baguan, a deep transformer model, we assume that the attention map functions similarly to the covariance matrix in a linear regression model. Based on this assumption, we perform spectral analyses on the attention scores associated with the center token in the map. Since there are a total of around 16,000 tokens, there are around 16,000 attention scores corresponding to this particular token. We observe that the energy of the leading eigenvectors is higher in the pre-trained model compared to the non-pre-trained model. We present the results in Fig.~\ref{fig:spectral_analysis}. Through this experiment, we aim to uncover the underlying mechanism of how pre-training operates, both from theoretical and empirical perspectives.

\subsubsection{Attention Map Visualization} Fig.~\ref{fig:intro}(c) illustrates the attention maps for the center token in the map for both the Siamese MAE and MAE in the $7^{\rm{th}}$ layer, within a total of 16 layers. 
The attention map of Siamese networks exhibits a stronger locality with lower-rank characteristics, effectively suppressing overfitting. This stronger locality implies that the attention map is more concentrated, with higher energy observed in the leading eigenvectors when viewed in the spectral domain. This phenomenon offers evidence for the conclusion in Section~\ref{sec:proof} that pre-training mitigates overfitting by pruning the subspace spanned by eigenvectors with small eigenvalues.

Specifically, when the attention values exceed a certain threshold, they are considered activated by the current token. In this context, \textbf{0.25\%} of tokens are activated by Siamese MAE, while \textbf{1.61\%} of tokens are activated more than six times by MAE.
Furthermore, the attention maps presented correspond to a 6-hour forecast, which is typically characterized by a relatively localized impact. However, MAE has activated nearly all tokens along the entire equator, indicating an unusually broad response.
This discrepancy is more likely indicative of an overfitting issue of MAE rather than an actual long-term effect on weather forecasting. A more detailed visualization of the attention maps can be found in Appendix~\ref{app:attn_map}.

\subsubsection{Ablation on Masking Ratio}

We conduct experiments to compare the effects of task difficulty by varying different masking ratios using the $1.40625^\circ$ resolution data for 6-hour prediction. It is important to note that a higher masking ratio corresponds to a more challenging pre-training task, making it increasingly similar to the w/o pre-training task. As shown in Fig. \ref{fig:ablation} (Right), the Siamese MAE with masking ratios of 0.75, 0.95, and 0.99 achieves average improvements of \textbf{8.4\%}, \textbf{8.8\%}, and \textbf{8.4\%} in RMSE on key variables, respectively, compared to the w/o pre-training version. While we observe optimal performance at a masking ratio of 0.95, increasing the masking ratio, particularly at 0.95 and 0.99, the pre-training process becomes unstable, characterized by significant spikes in the training loss. 
Consequently, to make a stable pre-training performance, we tend to choose a relative smaller masking ratio (e.g., 0.75) in our experiment.

\begin{figure*}[t]
    \centering
    \includegraphics[width=1\textwidth]{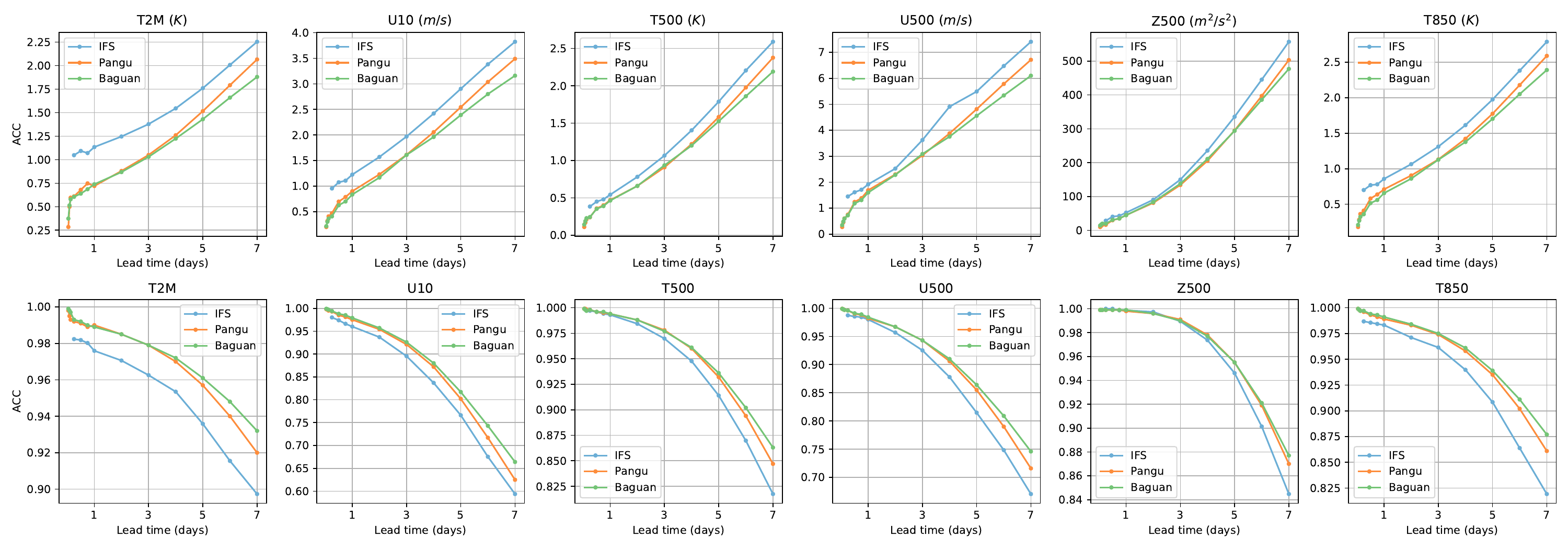}
    \caption{\small Comparison of global latitude-weighted RMSE and ACC against forecast lead time for \NAME (\textcolor{green}{green}), Pangu-Weather (\textcolor{orange}{orange}) and IFS (\textcolor{blue}{blue}). Key variables analyzed include T2M, U10, T500, U500, Z500 and T850. More detailed results can be found in Appendix~\ref{app:main_res}.}
    \label{fig:res_main_rmse}
\end{figure*}

\subsection{Weather Forecasting Tasks on $0.25^\circ$ ERA5 Dataset}

In Fig.~\ref{fig:res_main_rmse}, we present a comprehensive evaluation of \NAME (depicted by green lines), in comparison with Pangu-Weather and IFS models. The test set utilizes data from the year 2018 at 00 and 12 UTC. We compare the RMSE and ACC across lead times ranging from 1 hour to 7 days. 
\NAME demonstrates a lower RMSE than Pangu-Weather for \textbf{75.2\%} of the key variables, with an average improvement of \textbf{4.7\%}. Notably, for key variables such as T2M and U10, \NAME outperforms Pangu-Weather with average improvements of \textbf{4.0\%} and \textbf{7.8\%}, respectively

Additionally, \NAME consistently outperforms IFS, achieving an average improvement of \textbf{23.2\%}. In particular, for T2M, \NAME results in an average enhancement of \textbf{28.3\%}. As the lead time increases, the magnitude of our improvement becomes even more evident.

Our goal is to thoroughly study the effectiveness of pre-training using a simple ViT-based architecture without model-specific locality design. We did not compare our results with models~\cite{Fuxi_nature,graphcast,Chen2023FengWuPT}, as they mainly use GNNs or local attention mechanisms to induce locality and also we lack training codes for these algorithms to make a fair comparison. Future work will explore combining model-specific locality with pre-training-induced locality to collaboratively address the overfitting issue to make more accurate forecast.




\subsection{\NAME as Foundation Model}

\subsubsection{\NAME for S2S}

\begin{figure}[h]
    \centering
\includegraphics[width=1\columnwidth]{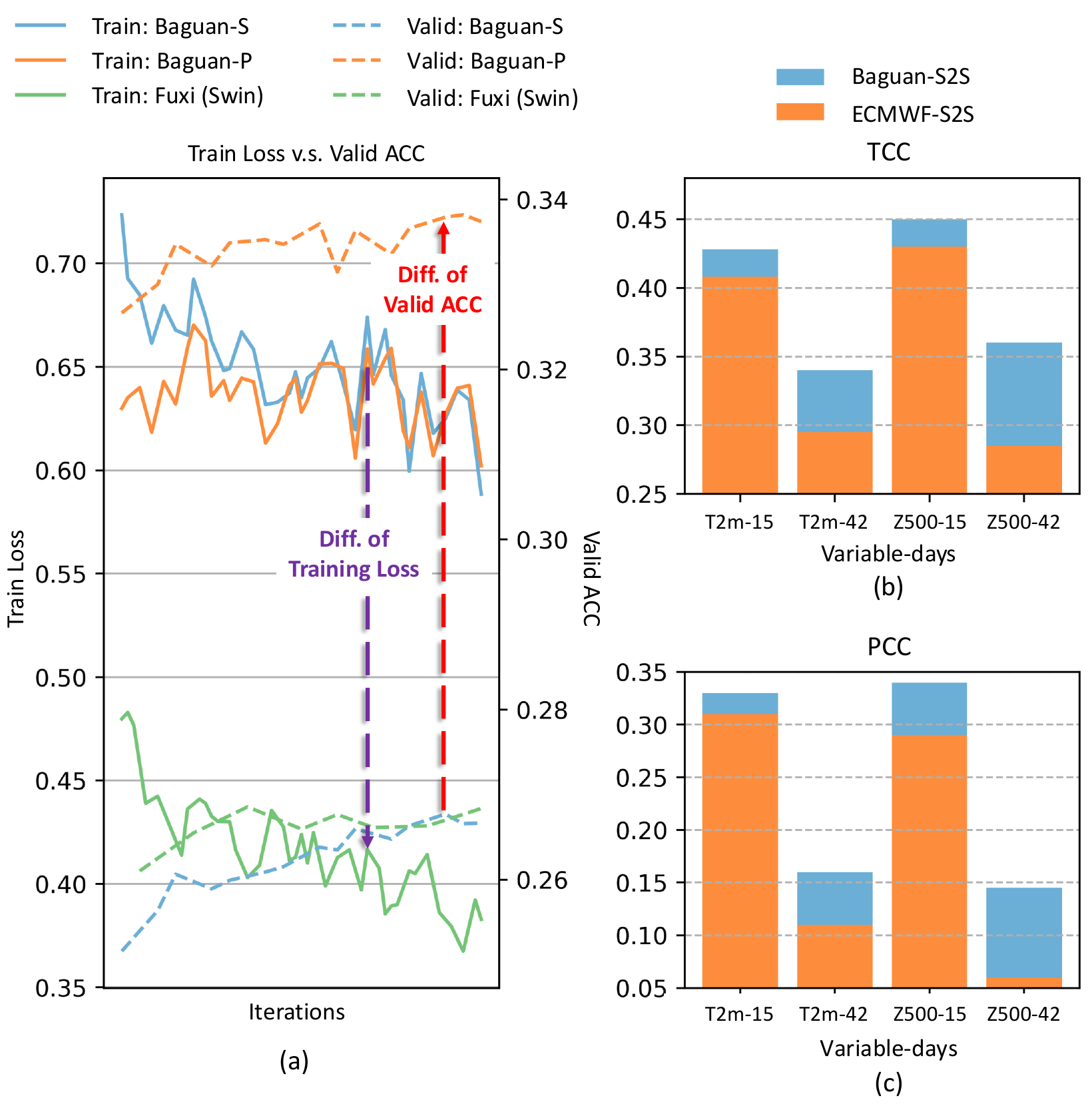}
    \caption{\small(a) The train loss and valid ACC of \NAME-S, \NAME-P and FuXi (swin-based model) on S2S task. \textcolor{black}{*-S represents train from sratch and *-P represents pre-training}. (b-c) The TCC and PCC examined for \NAME-S2S and ECMWF-S2S models in forecasting T2M and Z500. \textcolor{black}{*-15 and *-24 indicate a lead time of 15 and 42 days, respectively}. higher values of both TCC and PCC indicate better performance. The detailed results can be found in \cite{guo2024maximizing}.}
    \label{fig:s2s}
\end{figure}

Weather and climate forecasting covers a wide range of timescales. In addition to predicting weather conditions within the medium-range, subseasonal to seasonal (S2S) forecasting (2 to 6 weeks ahead)~\cite{subseasonal2seasonalprediction} also holds significant value in terms of disaster preparedness, such as droughts and floods. 
By fine-tuning \NAME with a novel multi-stage optimization approach, \cite{guo2024maximizing} demonstrates an impressive achievement in the S2S task. Their results outperform SOTA NWP systems(ECMWF-S2S), as illustrated in Fig. \ref{fig:s2s}. Recent benchmark work~\cite{nathaniel2024chaosbench} indicates that most deep learning models have shown weaker performance compared to ECMWF-S2S. 

Additionally, Fig. \ref{fig:s2s}(a) presents an analysis of overfitting among \NAME-P (pre-training), \NAME-S (train from scratch), and FuXi, a model based on the Swin Transformer architecture. Despite observing a low training loss with FuXi (0.38) and a higher training loss with \NAME (0.60), the \NAME-P model achieves significantly better validation accuracy, improving the TCC value by over \textbf{25\%}, from 0.265 to 0.337, compared to its w/o pre-training counterpart. This underscores \NAME-P's capability to mitigate overfitting through pre-training. We argue that in long-term subseasonal forecasting, where the signal-to-noise ratio is low, the risk of overfitting becomes more severe. In this context, \NAME's ability to address overfitting is particularly evident.

\subsubsection{\NAME for Regional Forecasting}

Regional forecasting achieving kilometer-scale predictions is vital for applications such as risk assessment and understanding local effects~\cite{NvidiaDiffusionForKmScaleWeather}. To enhance the precision of these forecasts, we have advanced beyond the traditional $0.25^\circ$ resolution of the ERA5 dataset by introducing a finer regional dataset with a resolution of $0.05^\circ$. By incorporating an additional encoder to process the tokens from this high-resolution regional analysis data and integrating them with global tokens, 
\NAME outperforms simple interpolation to high resolution. The forecasting system has been deployed to assist the electricity sector in an eastern province of China, covering over 1.5 million square kilometers and impacting 100 million people.

We perform short-term regional forecasting experiments utilizing data from the China Meteorological Administration Land Data Assimilation System (CLDAS). This dataset features $0.05^\circ$ grid spacing across China (16-54°N, 74-134°E) for the years 2022 and 2023, using the first one and a half years for training and the remaining half year for testing.
As results shown in Table~\ref{tab:regional_forecast_main}, we compare the regional forecasting outcomes of EC-IFS, Baguan-Interpolation (which interpolates the forecasts from the foundation model), and Baguan-Regional (which includes an additional encoder for regional analysis data). Baguan-Regional demonstrates significantly superior performance compared to EC-IFS, achieving an average improvement of \textbf{57.8\%} on key variables in regional forecasting. \NAME can be fine-tuned on variables not present in the pre-trained dataset, making it highly suitable for real-world applications. Additionally,it outperforms naive interpolation by an average of \textbf{47.0\%} on the same variables. These results highlight \NAME's enhanced forecasting capabilities at higher resolutions, achieved with only minimal adjustments. A more comprehensive description and detailed results can be found in Appendix~\ref{appendix:regional}.

\begin{table}[h]
\centering
\vskip -0.1in
\caption{Comparison of RMSE Between EC-IFS, Baguan-Interpolation, and Baguan-Regional.}
\vskip -0.1in
\label{tab:regional_forecast_main}
\begin{tabular}{cccccc}
\toprule
 \multirow{2}{*}{Variable} & Lead & \multirow{2}{*}{EC-IFS} & \NAME- & \NAME- \\
 &time&& Interpolation & Reigional \\
 \midrule
 \multirow{2}{*}{$T2m$} & 6h & 2.04  & 2.29 & \textbf{1.38} &  \\
 & 24h & 2.05  & 2.36 & \textbf{1.63} &  \\
 \midrule
 \multirow{2}{*}{$U10$} & 6h &  4.51 & 3.95 & \textbf{1.33} &  \\
 & 24h & 4.82  & 4.60 & \textbf{2.26} &  \\
\bottomrule
\end{tabular}
\end{table}
\vskip -0.1in


\subsection{Pre-Training Across Diverse Scales}

\subsubsection{Model Scaling Analysis}
In Fig.~\ref{fig:4_scaling_law} (Left), we evaluate the scaling laws for models with and without pre-training, revealing three key findings as follows. (1) Pre-training demonstrates greater effectiveness for larger models (\textbf{+4.20\% -> +10.1\%} at $5.625^\circ$ and \textbf{+6.02\% -> 8.03\%} at $1.40625^\circ$), suggesting that the benefits of pre-training scale with model capacity. (2) Model performance increases with model size, but saturates as the model reaches a sufficient scale. (3) Our analysis underscores the critical role of data resolution in model performance. Higher resolution inputs, which inherently contain more tokens, consistently lead to improved results. Notably, increasing the data resolution to $0.25^\circ$ reduces the validation MSE by \textbf{74.0\%} and \textbf{38.8\%} compared to resolutions of $5.625^\circ$ and $1.40625^\circ$, respectively.

\subsubsection{Data Scaling Analysis}

\begin{figure}[h]
    \centering
    \includegraphics[width=0.495\columnwidth]{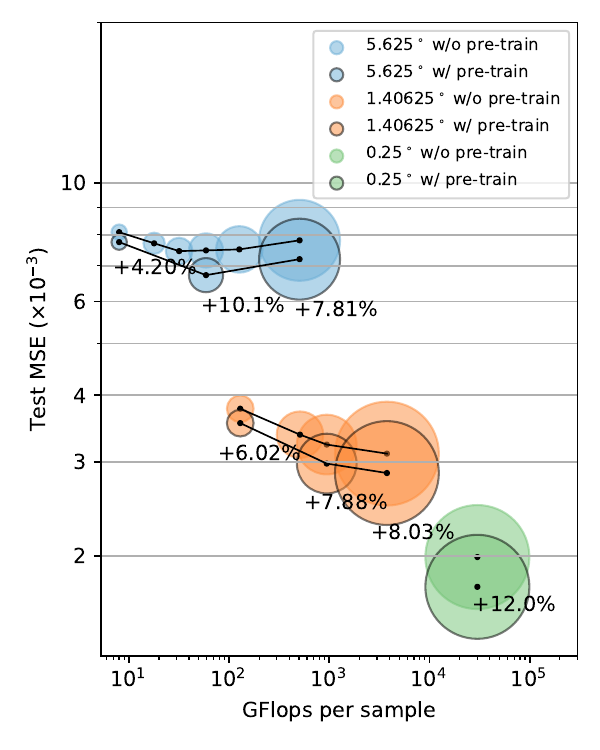}
    \includegraphics[width=0.495\columnwidth]{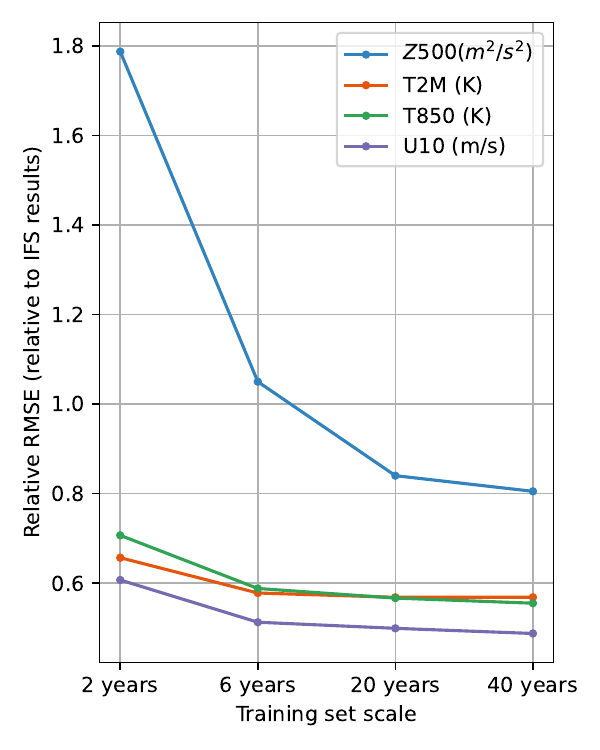} 
    \caption{\small (Left): the scaling analysis of model. We present an analysis of GFlops against test MSE for a series of models varying in size from 10M to 1300M, indicated by the size of the circles. Models trained at resolutions of $5.625^\circ$, $\1.40625^\circ$, and $\0.25^\circ$ are depicted in orange, green, and red, respectively. Models that are pre-trained are highlighted with black edges, whereas those without pre-training do not have black edges. (Right): the scaling analysis of data. We vary the training dataset scale from 40 years to 2 years and evaluated the performance on 4 key variables.}
    \label{fig:4_scaling_law}
\end{figure}

Fig.~\ref{fig:4_scaling_law} (Right) illustrates the empirical relationship between data scale and model performance. As discussed in previous sections, the constrained temporal scale of the ERA5 dataset (40 years) presents a fundamental limitation, resulting in model overfitting. Our experimental results substantiate this finding: while training loss consistently converges to similar levels across different dataset sizes, validation loss deteriorate with reduced data volume. This suggests that smaller datasets, due to their limited diversity in representing Earth's atmospheric states, consistently result in poorer generalization performance.The utilization of a 40-year dataset demonstrates average improvements of \textbf{27.4\%}, \textbf{8.9\%}, and \textbf{2.1\%} over datasets spanning 2 years, 6 years, and 20 years, respectively. Furthermore, among the four key variables analyzed, z500 exhibits the highest susceptibility to overfitting compared to the other three variables.

\subsection{Visualization}
\begin{figure}[h]
    \centering
    \includegraphics[width=0.97\columnwidth]{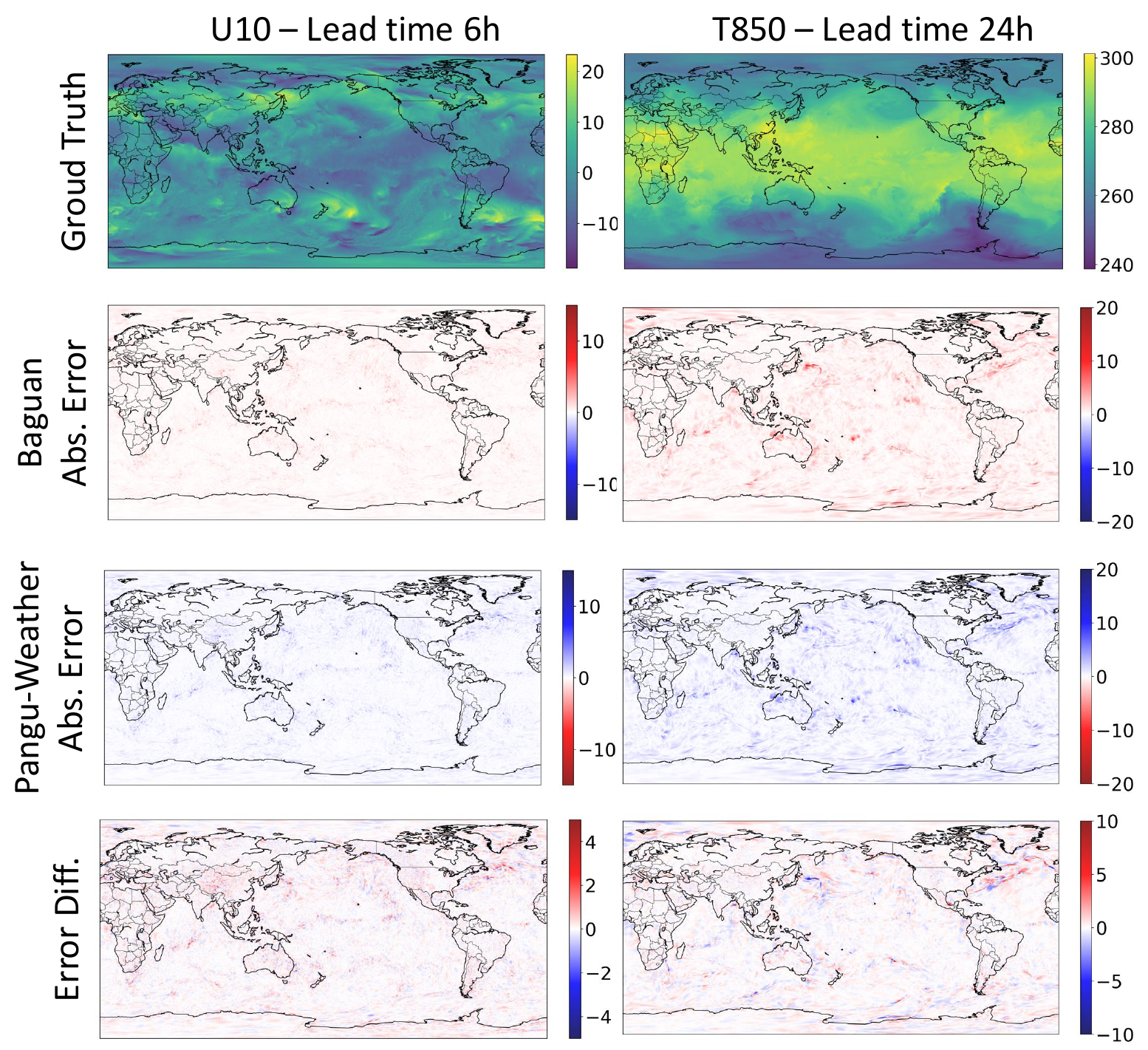}
    \caption{\small Visualization of absolute error. The spatial map showcases the absolute error for \NAME (second row) and Pangu-Weather (third row), as well as the difference in absolute error between \NAME and Pangu-Weather (fourth row) at forecast lead times of 6 hours and 24 hours. In the fourth row, \textcolor{red}{Red} represents \NAME performs better and \textcolor{blue}{Blue} represents Pangu-Weather performs better. The key variables analyzed include T850 and U10, with an initialization time of January 1, 2018, at 00:00 UTC. More detailed results can be found in Appendix~\ref{app:vis}.}
    \label{fig:vis}
\end{figure}

Fig.~\ref{fig:vis} illustrates the spatial distributions of absolute error for \NAME and Pangu-Weather at lead times of 6 hours and 24 hours. The initial state is set for January 1, 2018, at 00:00 UTC. 
The locations with discrepancies on the map are more apparent in Pangu-Weather.
Additionally, We visualize the difference in absolute error between \NAME and Pangu-Weather. The results indicate that a greater portion (\textbf{73\%} for U10 and \textbf{58\%} for T850) of the spatial map is red, suggesting that our \NAME model performs better.


\section{Conclusion and Future Work}


\paragraph{Conclusion} In this paper, we introduce \NAME, an advanced data-driven system designed to push the boundaries of weather forecasting. Our work delivers three major technical contributions: (1) We conduct a systematic investigation into pre-training techniques to effectively mitigate overfitting and enhancing the model's generalizability and robustness. (2) \NAME demonstrates superior performance, outperforming IFS and Pangu-Weather across a range of benchmarks.
(3) We show that \NAME excels in various downstream tasks, including S2S and regional forecasting, showcasing its versatility and applicability across different temporal and spatial scales in meteorological prediction. 

\paragraph{Future Work} In the future, we will continue to explore and conduct research from multiple perspectives, including: (1) We aim to investigate larger model architectures, employing advanced pre-training techniques to enhance the model's robustness and performance. 
(2) For downstream tasks, we aim to apply \NAME to a broader range of meteorological applications, including extreme weather forecasting and climate prediction. Appendix.~\ref{tracking_tc} presents an initial attempt at tracking tropical cyclones using Baguan. (3) To meet the diverse needs of our users, we will develop a modular system allowing users to select specific functional components and provide more flexible deployment options, ensuring better adaptability to different operational environments. 



\newpage
\bibliographystyle{ACM-Reference-Format}
\bibliography{sample-base,iclr2025_conference}

\onecolumn 

\appendix
\section{Appendix}

\subsection{Dataset}
\label{app:datset}
As illustrated in Table~\ref{sample-table}, we employ a subset of 13 pressure levels (50 hPa, 100 hPa, 150 hPa, 200 hPa, 250 hPa, 300 hPa, 400 hPa, 500 hPa, 600 hPa, 700 hPa, 850 hPa, 925 hPa, and 1000 hPa), along with four surface variables (2-meter temperature, u-component and v-component of 10-meter wind speed, and mean sea level pressure), as well as several statistical variables.

\begin{table*}[h]
\caption{ECMWF variables used in the training of \NAME.}
\label{sample-table}
\begin{center}

\resizebox{0.9\textwidth}{!}{
\begin{tabular}{llccccc}
\toprule
Type & Variable name & Abbrev. & Input & Output & ECMWF ID & Levels \\
\midrule
Static & Land-sea mask & LSM & $\checkmark$ & $\times$ & 172 & -\\
Static & geopotential at surface & - & $\checkmark$ & $\times$ & 162051 & -\\
Static & angle of sub gridscale orography & ANOR & $\checkmark$ & $\times$ & 162  & -\\
Static & anisotropy of sub gridscale orography & ISOR & $\checkmark$ & $\times$ & 161 & -\\
Static & lake cover & CL & $\checkmark$ & $\times$& 26 & -\\
Static & soil type & SOILTYP & $\checkmark$ & $\times$ & 500205 & - \\
\midrule
Single & 2 metre temperature & T2m & $\checkmark$ & $\checkmark$ & 167 & - \\
Single & 10 metre U wind component & U10 & $\checkmark$ & $\checkmark$& 165 & -\\
Single & 10 metre V wind component & V10 & $\checkmark$ & $\checkmark$& 166 & -\\
Single & mean sea-level pressure & MSL & $\checkmark$ & $\checkmark$& 151 & -\\
\midrule
\multirow{2}{*}{Atmospheric} & \multirow{2}{*}{Geopotential} & \multirow{2}{*}{Z} & \multirow{2}{*}{$\checkmark$} & \multirow{2}{*}{$\checkmark$} & \multirow{2}{*}{129}
& 50, 100, 150, 200, 250, 300, \\
&&&&&& 400, 500, 600, 700, 850, 925, 1000 \\
\multirow{2}{*}{Atmospheric} & \multirow{2}{*}{U wind component} & \multirow{2}{*}{U} & \multirow{2}{*}{$\checkmark$} & \multirow{2}{*}{$\checkmark$} & \multirow{2}{*}{131}
& 50, 100, 150, 200, 250, 300, \\
&&&&&& 400, 500, 600, 700, 850, 925, 1000 \\
\multirow{2}{*}{Atmospheric} & \multirow{2}{*}{V wind component} & \multirow{2}{*}{V} & \multirow{2}{*}{$\checkmark$} & \multirow{2}{*}{$\checkmark$} & \multirow{2}{*}{132}
& 50, 100, 150, 200, 250, 300, \\
&&&&&& 400, 500, 600, 700, 850, 925, 1000 \\
\multirow{2}{*}{Atmospheric} & \multirow{2}{*}{Temperature} & \multirow{2}{*}{T} & \multirow{2}{*}{$\checkmark$} & \multirow{2}{*}{$\checkmark$} & \multirow{2}{*}{130}
& 50, 100, 150, 200, 250, 300, \\
&&&&&& 400, 500, 600, 700, 850, 925, 1000 \\
\multirow{2}{*}{Atmospheric} & \multirow{2}{*}{Specific humidity} & \multirow{2}{*}{Q} & \multirow{2}{*}{$\checkmark$} & \multirow{2}{*}{$\checkmark$} & \multirow{2}{*}{133}
& 50, 100, 150, 200, 250, 300, \\
&&&&&& 400, 500, 600, 700, 850, 925, 1000 \\
\bottomrule
\end{tabular}
}
\end{center}
\end{table*}

The "Type" column categorizes variables as static (non-time-varying) properties, time-varying, single-level (surface) properties, or time-varying atmospheric properties. The "Variable Name" and "Abbrev." columns denote the labels assigned by ECMWF, whereas the "ECMWF ID" column specifies the numerical identifier allocated by the organization. The "Input" and "Output" columns indicate whether a variable functions as an input to or an output from \NAME. Lastly, the "Levels" column details the atmospheric levels included in the dataset.

\subsection{Evaluation Metrics \& Training Objectives}
\label{app:metric}

\paragraph{Evaluation metrics}
RMSE is a commonly utilized statistical metric in geospatial analysis and climate science to evaluate the precision of a model's predictions or estimates over various latitudinal ranges:
\begin{equation}
    \label{rmse_metric}
    RMSE = \frac{1}{N} \Sigma_{t=1}^N \sqrt{\frac{1}{H W}\Sigma_{i=1}^H \Sigma_{j=1}^W L(i)(\tilde{X}_{t, i, j} - X_{t, i, j})^2}.
\end{equation}
The latitude weighting factor $L(i)$ is used to account for the differences in surface area represented by various latitudes on a spherical Earth. It is given by:
\begin{equation}
    \label{lat_weight}
    L(i) = \frac{\mathrm{cos}(\mathrm{lat}(i))}{\frac{1}{H} \Sigma_{k=1}^H\mathrm{cos}(\mathrm{lat}(j))}.
\end{equation}

ACC is the spatial correlation between prediction anomalies $\tilde{X}^{'}$ relative to climatology and ground truth anomalies $X^{'}$ relative to climatology:
\begin{equation}
    \label{acc}
    ACC = \frac{\Sigma_{t, i, j}L(i)\tilde{X}^{'}_{t,i,j}X^{'}_{t,i,j}}{\sqrt{\Sigma_{t, i, j}L(i)\tilde{X}^{'2}_{t,i,j}X^{'2}_{t,i,j}}},
\end{equation}
\begin{equation}
    \label{acc_app}
    \tilde{X}^{'} = \tilde{X} - C, X^{'} = X- C,
\end{equation}
where climatology $C$ is the temporal mean of the ground truth data over the entire test set $C = \frac{1}{N}\Sigma_kX$.

\paragraph{Training objectives}
We use the Mean Square Error (MSE) loss for
pre-training and Mean Absolute Error (MAE) loss for fine-tuning. Given the large number of variables being predicted, we employ pressure-weighted loss to prioritize the weighting of variables near the surface. These weights are normalized to ensure their total sums to 1. Additionally, to align with the evaluation metrics, latitude-weighting $L(i)$ is also incorporated in the training objective.
\begin{equation}
    \label{mse}
    \mathcal{L}_{\rm{pre-train}}^{MSE}(\theta) = \frac{1}{VHW}\Sigma_{v=1}^V\Sigma_{i=1}^H\Sigma_{j=1}^W \omega(v)L(i)(\tilde{X}_{v,i,j} - X_{v,i,j})^2,
\end{equation}
\begin{equation}
    \label{mae}
    \mathcal{L}_{\rm{fine-tune}}^{MAE}(\theta) = \frac{1}{VHW}\Sigma_{v=1}^V\Sigma_{i=1}^H\Sigma_{j=1}^W \omega(v)L(i)|\tilde{X}_{v,i,j} - X_{v,i,j}|,
\end{equation}
where $\omega(v)$ is the weight of variable $v$.

\begin{table*}[t]
\centering
\caption{Model Configurations}
\begin{tabular}{c|cccccccccc}
\toprule
Config&\textsc{\NAME-$0.25^{\circ}$}&\textsc{\NAME-$1.40625^{\circ}$}&\textsc{\NAME-$5.625^{\circ}$} \\

\midrule
Encoder Layers &16&16&8 \\
Encoder Embed Dim &2048&1024&512 \\
Encoder Heads &16&16&16 \\
\midrule
Decoder Layers &4&4&4 \\
Decoder Embed Dim & 1024 & 512 & 256 \\
Decoder Heads &8&8&8 \\
\midrule
Patch Size &8&4&2 \\
Masking Ratio &0.75&0.75&0.75 \\
Scheduler &Cosine&Cosine&Cosine \\
Optimizer &AdamW&AdamW&AdamW \\
\midrule
Learning Rate - Stage 1 &5e-4& 4e-4 &2e-4 \\
Warm-up Steps - Stage 1 &40,000& 20,000  &10,000 \\
Max Steps - Stage 1 &250,000& 200,000 &100,000 \\
$\beta_1$ - Stage 1 &0.9&0.9&0.9 \\
$\beta_2$ - Stage 1 &0.95&0.95&0.95 \\
Weight Decay - Stage 1 &0.05&0.05&0.05 \\
Batch Size - Stage 1 & 24 & 32 & 64 \\
Batch Size / GPU - Stage 1 & 1 & 4 & 16 \\
\midrule
Learning Rate - Stage 2 &2e-5& 1e-5&1e-5 \\
Warm-up Steps - Stage 2 &10,000& 10,000 &10,000 \\
Max Steps - Stage 2 &70,000& 60,000 &40,000 \\
$\beta_1$ - Stage 2 &0.9&0.9&0.9 \\
$\beta_2$ - Stage 2 &0.99&0.99&0.99 \\
Weight Decay - Stage 2 &1e-5&1e-5&1e-5 \\
Batch Size - Stage 2 &8& 32&64 \\
Batch Size / GPU - Stage 2 & 1 & 4 & 16 \\
\midrule
Learning Rate - Stage 3 &2e-6&2e-7&2e-7 \\
Warm-up Steps - Stage 3 &0&0&0 \\
Max Steps - Stage 3 &2,000&2,000&1,000 \\
$\beta_1$ - Stage 3 &0.9&0.9&0.9 \\
$\beta_2$ - Stage 3 &0.99&0.99&0.99 \\
Weight Decay - Stage 3 &1e-5&1e-5&1e-5 \\
Batch Size - Stage 3 &8&32&64 \\
Batch Size / GPU - Stage 3 & 1 & 4 & 16 \\
\bottomrule
\end{tabular}\label{tab:model_details}
\end{table*}

\subsection{Implementation Details of Each Stage}
\label{app:imp}
\NAME undergoes three training stages: (1) pre-training on the Siamese Masked Image Modeling (SMIM) tasks, (2) fine-tuning for specific lead times of 1 hour, 6 hours, or 24 hours, and (3) iterative fine-tuning utilizing multi-step data.

The \textbf{Stage 1} for pre-training requires about 14 days on a cluster of 24 Nvidia A800 GPUs. The model undergoes a warm-up process in which the learning rate is increased linearly to 5e-4, and then gradually annealed to 0 following a cosine schedule with a total of 250,000 steps. The AdamW optimizer is used with parameters $\beta_1 = 0.9$ and $\beta_2 = 0.95$, and a weight decay coefficient of 0.05. The masking ratio for siamese MAE is $75\%$.

The \textbf{Stage 2} involves the fine-tuning of the 24-hour, 6-hour and 1-hour models on 8 Nvidia A800 GPUs. This process requires about 4 days for each model, totaling 70,000 steps.  The AdamW optimizer is used with parameters $\beta_1 = 0.9$ and $\beta_2 = 0.99$, an linearly increased learning rate of 2e-5, and a weight decay coefficient of 1e-5.

The \textbf{Stage 3} for iterative fine-tuning on 8 Nvidia A800 GPUs requires a total of 5 days. The AdamW optimizer is used with parameters $\beta_1 = 0.9$ and $\beta_2 = 0.99$, an fixed learning rate of 2e-6, and a weight decay coefficient of 1e-5. 
Similar to GraphCast \cite{graphcast}, the number of iterative steps is raised from 2 to 7 (24 hour model), 3 (6 hour model) or 5 (1 hour model), with an increment of 1 every 2000 steps.
However, due to the memory limits for longer steps in auto-regressive paradigm, we simply perform back propagation for each individual step.

\subsection{Model Configuration}
\label{app:model_config}
Table~\ref{tab:model_details} provides the detailed model configurations of \NAME on $0.25^{\circ}$, $1.40625^{\circ}$ and $5.625^{\circ}$ weather forecasting tasks. Overall, the model configurations are designed to balance complexity and compute resource demands with changing forecasting scales. Higher resolution models require more computational resources and longer training steps, while lower resolution models are more lightweight.

\subsection{Ablations of Ensemble Strategy}
\label{app:ensemble}

Table~\ref{tab:ensemble} provides the ablation of our ensemble strategy. Equipped with this strategy, \NAME can achieve an average improvement of approximately 2\% in forecasting at longer lead times.
\begin{table}[h]
\centering
\caption{Ablations of Ensemble Strategy. Lower RMSE represents better performance.}
\label{tab:ensemble}
\begin{tabular}{c|ccccc|ccccc}

\toprule
 & \multicolumn{5}{c|}{48h} & \multicolumn{5}{c}{72h} \\
& T2m & U10 & MSL & T850 & Z850 & T2m & U10 & MSL & T850 & Z850 \\

\midrule

\NAME w/o Ensemble & 0.905 & 1.203 & 92.21 & 0.877 & 67.30 & 1.049 & 1.609 & 147.6 & 1.118 & 107.7 \\
\NAME w/ Ensemble & 0.868 & 1.167 & 88.94 & 0.860 & 65.19 & 1.050 & 1.600 & 145.7 & 1.126 & 106.4 \\

\bottomrule
\end{tabular}
\end{table}

\subsection{Detailed Results on $0.25^{\circ}$}
\label{app:main_res}
\begin{figure}[h]
    \centering
    \includegraphics[width=0.8\textwidth]{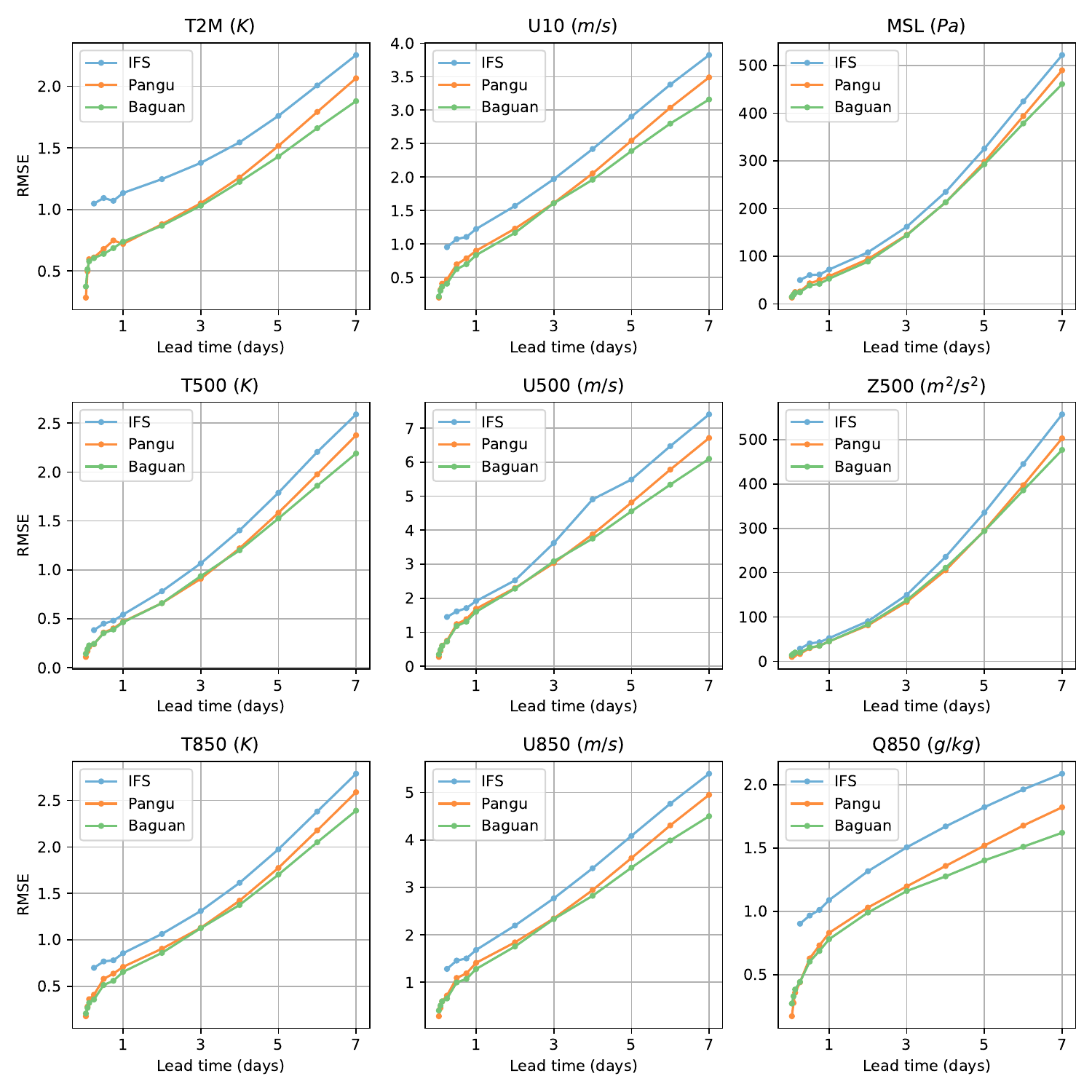}
    \caption{Comparison of global latitude-weighted RMSE of $0.25^{\circ}$ data for \NAME and Pangu-Weather.}
    \label{fig:main_res_app}
\end{figure}
Fig.~\ref{fig:main_res_app} provide a more detailed results on $0.25^{\circ}$ data. The key variables analyzed include T2M, U10, MSL, T500, U500, Z500, T850, U850 and Q850.


\subsection{Full Results of $1.40625^{\circ}$ and $5.625^{\circ}$}
\label{app:res_1_5}
\begin{figure}[h]
    \centering
    \includegraphics[width=1\textwidth]{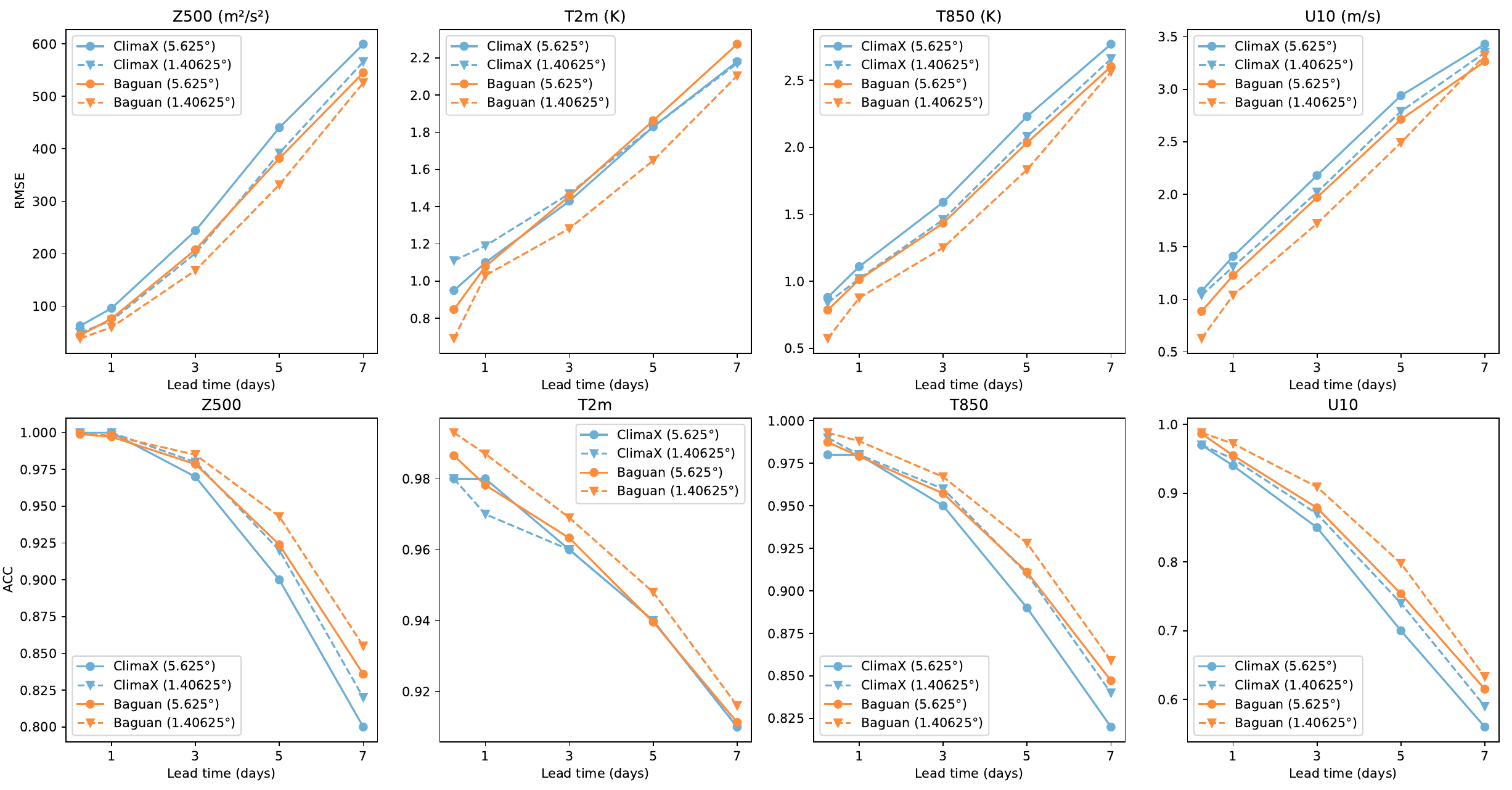}
    \caption{Results of $1.40625^{\circ}$ and $5.625^{\circ}$ for \NAME and ClimaX.}
    \label{fig:res_1_5}
\end{figure}

ClimaX \cite{Nguyen2023ClimaXAF} is pre-trained on CMIP data through directly forecasting. Fig.~\ref{fig:res_1_5} shows the comparison between \NAME and ClimaX on $1.40625^{\circ}$ and $5.625^{\circ}$. \NAME consistently outperforms ClimaX across all variables and lead times. To ensure a fair comparison, we employ a \NAME model with a size similar to ClimaX, featuring a hidden dimension of 1024 and a depth of 8. The experimental setup in this study adheres to the guidelines outlined in \cite{Nguyen2023ClimaXAF}.

\newpage
\subsection{Attention Map Across Various Methods}
\label{app:attn_map}
\begin{figure}[h]
    \centering
    \includegraphics[width=1\textwidth]{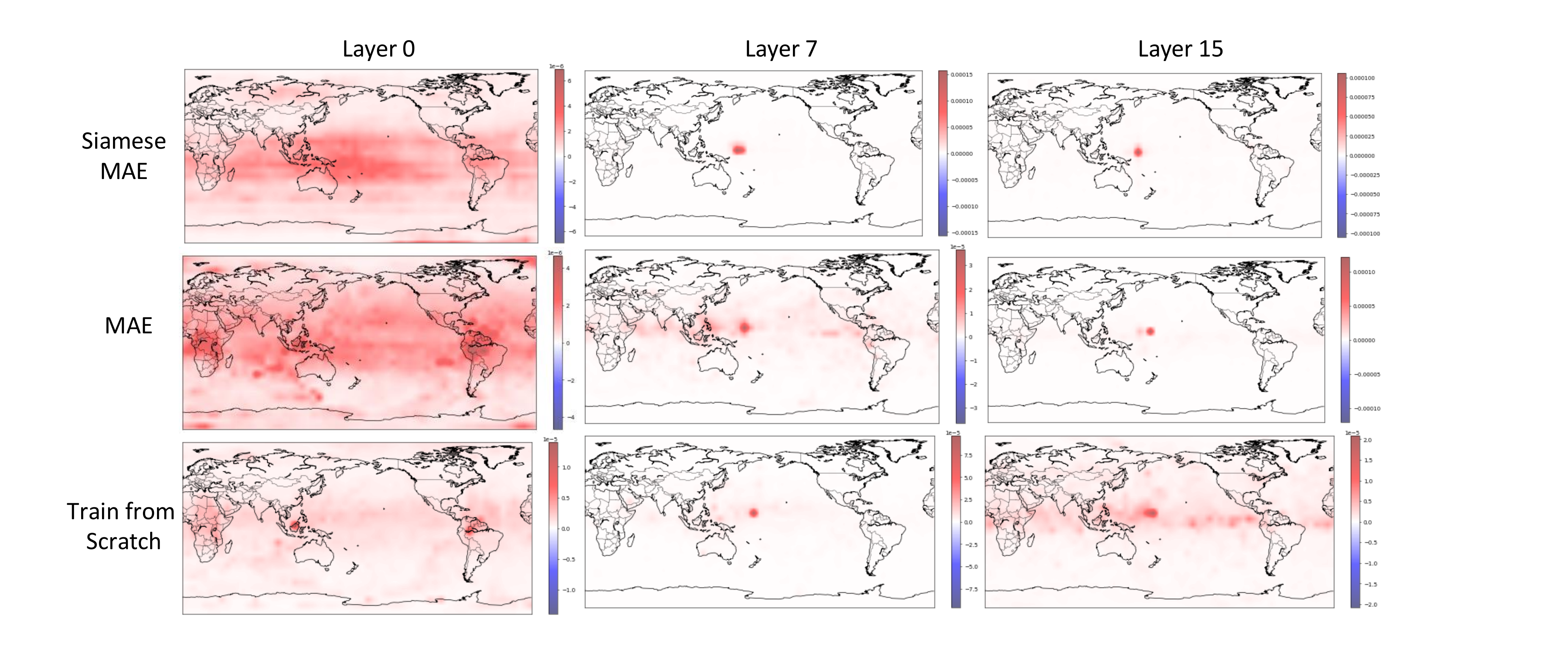}
    \caption{Detailed attention map of Siamese MAE, MAE and the model trained from scratch.}
    \label{fig:attn_map_app}
\end{figure}

Fig.~\ref{fig:attn_map_app} provides a comparative visualization of encoder attention maps during the inference phase for Siamese MAE, MAE, and the model trained from scratch. As demonstrated by \NAME, the ideal attention map progressively shifts from global to local patterns as transformer layers deepen, forming a pyramidal structure that reflects hierarchical feature learning.

\newpage

\subsection{Visualization}
\label{app:vis}
\begin{figure}[t]
    \centering
    \includegraphics[width=1\textwidth]{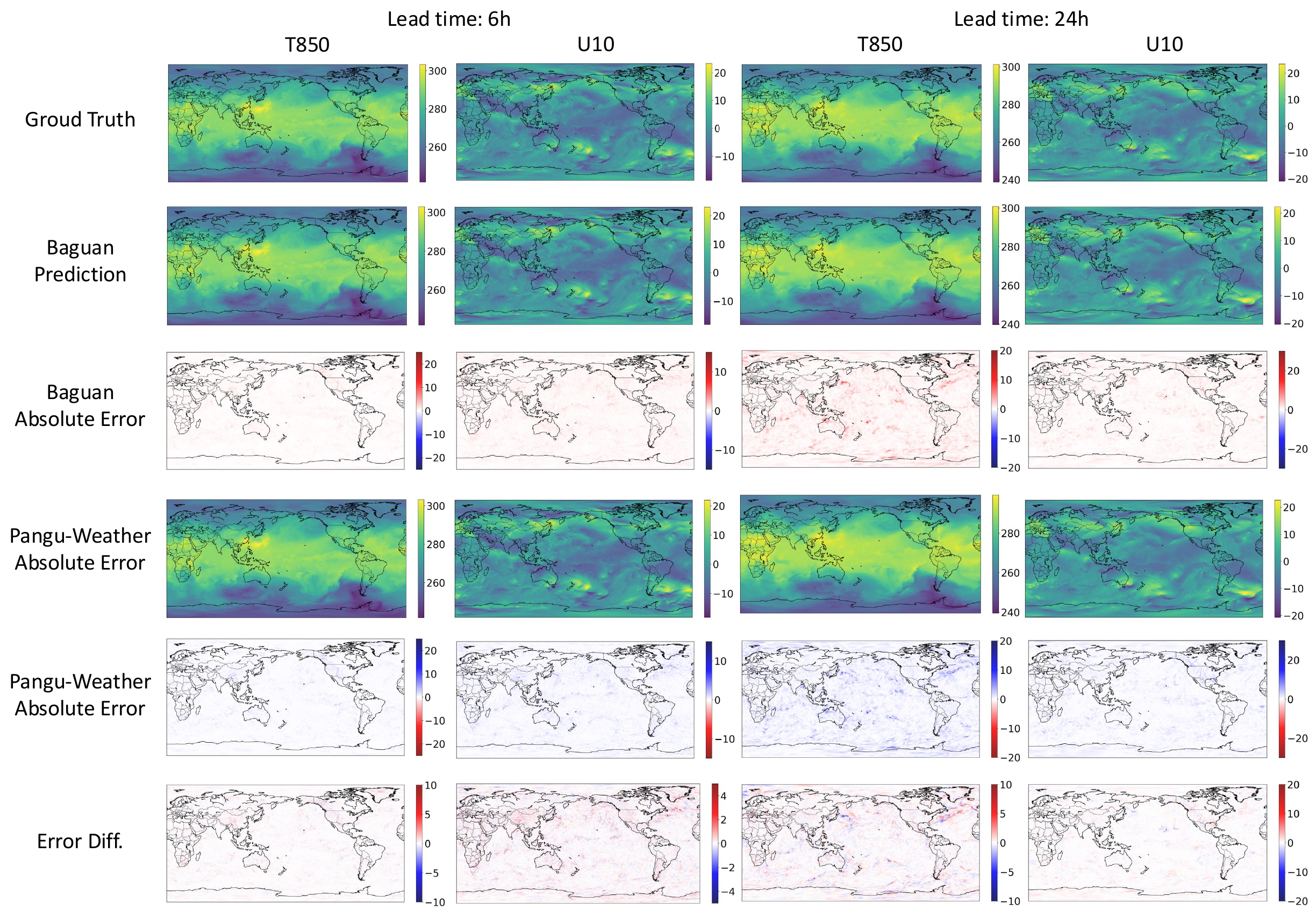}
    \caption{Detailed visualization of \NAME and Pangu-Weather.}
    \label{fig:vis_app}
\end{figure}

Fig.~\ref{fig:vis_app} presents a comparative visualization of the ground truth atmospheric state, \NAME-predicted values, Pangu-Weather predictions, and their corresponding error distributions. The figure highlights errors for both models (\NAME and Pangu-Weather) and quantifies their performance gap by explicitly visualizing the difference between \NAME and Pangu-Weather prediction errors, providing direct insight into model discrepancies.

\subsection{Comparison of Baseline for Baguan's regional weather forecasting task}
\label{appendix:regional}
\begin{table}[h]
\centering
\caption{Comparison of RMSE Between EC-IFS, Baguan-Interpolation, and Baguan-Regional}\label{tab:regional_forecast}
\renewcommand{\arraystretch}{0.6}
\setlength\tabcolsep{7pt}
\resizebox{0.6\columnwidth}{!}{
\begin{tabular}{cccccc}
\toprule
\specialrule{0em}{0.5pt}{0.5pt}
{\scalebox{0.85}{Variable}} & \scalebox{0.85}{Lead-Time} & \scalebox{0.9}{EC-IFS} & \scalebox{0.9}{Baguan-Interpolation} & \scalebox{0.9}{Baguan-Regional} &  \\
 \specialrule{0em}{0.5pt}{0.5pt} \Xhline{0.5px} \specialrule{0em}{0.5pt}{0.5pt}
 \multirow{4}{*}{$T2m$} & 6 & 2.04  & 2.29 & \textbf{1.38} &  \\
 & 12 & 2.04  & 2.31 & \textbf{1.45} & \\
 & 18 & 2.12  & 2.37 & \textbf{1.52} &  \\
 & 24 & 2.05  & 2.36 & \textbf{1.63} &  \\
 \specialrule{0em}{0.5pt}{0.5pt} \Xhline{0.5px} \specialrule{0em}{0.5pt}{0.5pt}
 \multirow{4}{*}{$U10$} & 6 &  4.51 & 3.95 & \textbf{1.33} &  \\
 & 12 & 4.60  & 4.23 & \textbf{1.69} & \\
 & 18 & 4.90  & 4.45 & \textbf{2.01} &  \\
 & 24 & 4.82  & 4.60 & \textbf{2.26} &  \\
 \specialrule{0em}{0.5pt}{0.5pt} \Xhline{0.5px} \specialrule{0em}{0.5pt}{0.5pt}
 \multirow{4}{*}{$V10$} & 6 & 4.68  & 4.01 & \textbf{1.35} &  \\
 & 12 & 4.71  & 4.27 & \textbf{1.77} & \\
 & 18 & 5.03  & 4.53 & \textbf{2.08} &  \\
 & 24 & 4.86  & 4.74 & \textbf{2.28} &  \\
 \specialrule{0em}{0.5pt}{0.5pt} \Xhline{0.5px} \specialrule{0em}{0.5pt}{0.5pt}
 \multirow{4}{*}{$SP$} & 6 &  16.23 & - & \textbf{6.30} &  \\
 & 12 & 16.28  & - & \textbf{7.23} & \\
 & 18 & 15.94  & - & \textbf{7.95} &  \\
 & 24 & 16.11  & - & \textbf{8.24} &  \\
\specialrule{0em}{0.5pt}{0.5pt} \Xhline{0.5px} \specialrule{0em}{0.5pt}{0.5pt}
 \multirow{4}{*}{$SH$} & 6 & 8.42e-4  & - & \textbf{7.89e-4} &  \\
 & 12 & \textbf{7.78e-4}  & - & 8.28e-4 & \\
 & 18 & 8.63e-4  & - & \textbf{8.61e-4} &  \\
 & 24 & 9.93e-4  & - & \textbf{9.10e-4} &  \\
\bottomrule
\end{tabular}}
\end{table}
Regional forecasting achieving kilometer-scale predictions is vital for applications such as risk assessment and understanding local effects~\cite{NvidiaDiffusionForKmScaleWeather}. To enhance the precision of these forecasts, we have advanced beyond the traditional $0.25^\circ$ resolution of the ERA5 dataset by introducing a finer regional dataset with a resolution of $0.05^\circ$. By incorporating an additional encoder to process the tokens from this high-resolution regional data and integrating them with global tokens, 
\NAME outperforms interpolation forecastings to high resolution. 

We perform short-term regional forecasting experiments utilizing data from the China Meteorological Administration Land Data Assimilation System (CLDAS). This dataset features $0.05^\circ$ grid spacing across China (16-54°N, 74-134°E) for the years 2022 and 2023, using the first one and a half years for training and the remaining half year for testing. The forecasted variables include 2-meter temperature (T2M), 10-meter zonal and meridional wind components (U10 and V10), surface pressure (SP), and specific humidity (SH). 

In Table \ref{tab:regional_forecast}, we compare the regional forecasting outcomes of EC-IFS, Baguan-Interpolation (which interpolates the forecasts from the foundation model), and Baguan-Regional (which includes an additional encoder for regional analysis data). Baguan-Regional significantly outperforms EC-IFS and interpolation methods in regional forecasting, illustrating \NAME's improved forecasting capabilities at higher resolutions with minimal adjustments.

\subsection{The Performance of Baguan in Tracking Tropical Cyclones}
\label{tracking_tc}

In our future work, we aim to thoroughly incorporate Baguan to address various extreme weather conditions. 
Here, we provide a initial attempt by directly using Baguan's forecasts for cyclone tracking. Similar to the approach taken in the Pangu paper, we identify the predicted typhoon position by selecting the point with the minimum mean sea level pressure (MSL) within a qualifying window, based on the initial position and starting time of the cyclone eye. The results in the Tab.~\ref{tab:tracking_tc} below demonstrate that Baguan outperforms both Pangu and IFS forecasts on average.

\begin{table}[h]
\centering
\caption{Tracking errors (km) of 5 typical cyclones in 2018. Lower is better.}
\label{tab:tracking_tc}
\begin{tabular}{c|ccc|ccc|ccc}

\toprule

 & \multicolumn{3}{c|}{TC1} & \multicolumn{3}{c|}{TC2} & \multicolumn{3}{c}{TC3} \\
Lead times (d)& Baguan & Pangu & IFS & Baguan & Pangu & IFS & Baguan & Pangu & IFS \\
\midrule
1	&55.59	&38.68	&137.33	&27.79	&27.79	&243.39	&27.79	&25.24	&75.17	\\
2	&61.68	&27.79	&136.22	&38.68	&61.75	&243.67	&104.63	&100.77	&80.46	\\
3	&98.95	&87.54	&212.67	&61.78	&55.59	&217.68	&27.79	&103.83	&103.83 \\
4	&52.95	&61.59	&202.86	&168.94	&149.04	&317.44	&73.80	&73.80	&150.36\\
5	&76.77	&143.91	&333.50	&223.99	&393.24	&283.66	&120.06	&100.11	&166.94 \\
\midrule

 & \multicolumn{3}{c|}{TC4} & \multicolumn{3}{c|}{TC5} & \multicolumn{3}{c}{Average} \\
 \midrule
Lead times (d) & Baguan & Pangu & IFS & Baguan & Pangu & IFS & Baguan & Pangu & IFS \\
1 &26.68	&26.68	&110.37	&26.97	&38.72	&121.47	&32.964	&\textbf{31.422}	&137.546 \\
2 &76.92	&97.21	&136.05	&26.68	&80.06	&144.71	&\textbf{61.718}	&73.516	&148.222 \\
3 &79.08	&76.56	&191.87	&55.59	&55.59	&106.04	&\textbf{64.638}	&75.822	&166.418 \\
4 &104.65 &78.49	&142.32	&83.43	&133.88	&185.66	&\textbf{96.754}	&99.36	&199.728\\
5 	&141.75	&118.21	&82.97	&140.76	&208.23	&284.01	&\textbf{140.666}	&192.74	&230.216\\


\bottomrule
\end{tabular}
\end{table}

\subsection{Pre-training as Regularization}
\label{appendix_provement}
To understand how pre-training helps mitigate the overfitting problem, we analyze the case of linear regression. We show that pretraining helps prune the subspace spanned by the eigenvectors for small eigenvalues, and as a result, regularizes the final solution by restricting the solution mostly to the top eigenvectors of covariance matrix. This regularization effect will help significantly reduce the overfitting problem when the target solution also lies in the subspace of top eigenvectors of covariance matrix. Below, we will first describe the linear regression problem, and state the result of the denoising based pre-training. We will then show how a denoising based pre-training regularizes the solution and improves the generalization performance

Let $(x_i, y_i), i=1, \ldots, n$ be a set of training examples, where $x_i \in \R^d$ and $y_i \in \R$, with $\E[x_ix_i^{\top}] = \Sigma$. Let $(\lambda_i, v_i), i=1, \ldots, d$ be the eigenvalue and eigenvector of $\Sigma$ ranked in the descending order. We assume that $y_i = w_*^{\top}x_i + z_i$, where $w_* \in \R^d$ is the optimal solution and $z_i \sim \N(0, \sigma^2)$ is a random noise sampled from a Gaussian distribution. We further assume that $w_*$ lies in the subspace spanned by the $K$ eigenvectors of $\Sigma$, i.e. $w_* = P_K w_*$, where $P_k = \sum_{i=1}^K v_i v_i^{\top}$ is a projection operator that projects any input vector into the subspace spanned by $v_1, \ldots, v_K$. We assume that $K \ll d$. 

Using the closed form solution for the ridged linear regression, we have the solution without pre-training as
\begin{eqnarray*}
w_1 & = & \left(\sum_{i=1}^n x_i x_i^{\top} + \lambda\right)^{-1}\sum_{i=1}^n x_iy_i \\
& = & \left(\sum_{i=1}^n x_i x_i^{\top} + \lambda\right)^{-1} \left(\sum_{i=1}^n x_i x_i^{\top}\right)w_* + \sum_{i=1}^n  \left(\sum_{i=1}^n x_i x_i^{\top}+\lambda I\right)^{-1}x_i z_i \\
& = & \left(\sum_{i=1}^n P_K x_i x_i^{\top} P_K + \lambda\right)^{-1} \left(\sum_{i=1}^n P_K x_i x_i^{\top}\right)P_K w_* + \sum_{i=1}^n  \left(\sum_{i=1}^n x_i x_i^{\top}+\lambda I\right)^{-1}x_i z_i
\end{eqnarray*}
where the last step leverages the fact that $w_* = P_K w_*$. 

Using matrix chernoff bound, we have, with a probability $1 - \delta$, that
\[
\lambda_{\min}\left(\frac{1}{n}\sum_{i=1}^n P_K x_i x_i^{\top} P_K\right) \geq \left(1 - \sqrt{\frac{3R}{n\lambda_K}\log\frac{1}{\delta}}\right)\lambda_K
\]
where $|x_i| \leq R$ for all $i \in [n]$. Hence, when
\[
n \geq \frac{12R}{\lambda_K}\log\frac{1}{\delta}
\]
we have, with a probability at least $1 - \delta$, 
\[
\lambda_{\min}\left(\frac{1}{n}\sum_{i=1}^n P_K x_i x_i^{\top} P_K\right) \geq \frac{\lambda_K}{2}
\]
Thus, we have, with a probability at least $1 - \delta$,
\[
|w_1 - w_*| \leq \frac{\lambda}{\lambda_K/2 + \lambda} |w_*| + \frac{1}{\lambda}\left|\frac{1}{n}\sum_{i=1}^n x_i z_i\right| 
\]
Using the concentration inequality for random vectors, we have, with a probability at least $1 - \delta$, 
\[
 \frac{1}{\lambda}\left|\frac{1}{n}\sum_{i=1}^n x_i z_i\right| \leq \frac{R}{n} + \sqrt{\frac{\sigma^2\mbox{tr}(\Sigma)}{n}\log\frac{1}{\delta}}
\]
Combining the above result, we have, with a probability $1 - 2\delta$
\begin{eqnarray}
|w_1 - w_*| \leq \frac{\lambda}{\lambda_K/2 + \lambda} |w_*| + \frac{R}{n} + \sqrt{\frac{\sigma^2\mbox{tr}(\Sigma)}{n}\log\frac{1}{\delta}} \label{eqn:bound-1}
\end{eqnarray}

We now study the case of ridge linear regression with pre-training. In particular, our pre-training is a denoising based algorithm, i.e. it searches for a matrix $M \in \R^{d\times d}$ such as we can recover $x_i$ from $x_i + \delta_i$, a noisy version of $x_i$ with $\delta_i \in \N(0, \gamma^)$. It is essentially cast into the following optimization problem
\[
\min\limits_{M \in \R^{d\times d}} \; \E_{x, \delta}\left[|M(x+\delta) - x|^2\right]
\]
Here we assume that we have enough number of unlabeled examples such that we simply express the objective function as a true expectation over both $x$ and $\delta$. The resulting optimal solution for $M$ is given by
\[
M_* = \Sigma\left(\Sigma + \gamma I\right)^{-1} = \sum_{i=1}^d \frac{\lambda_i}{\lambda_i + \gamma} v_i v_i^{\top}
\]
As indicated by the expression of $M_*$, its eigenvalue is close to $1$ when $\lambda_i \gg \gamma$, but becomes zero when $\lambda_i \ll \gamma$. Hence, we can view $M_*$ as a high bandwidth pass filter that filters out any vector lying the subspace of the eigenvectors for small eigenvalues. With the pre-trained matrix $M_*$, we will build a ridge regression model on top of it, i.e. transform each input $x_i$ into $M_* x_i$. The resulting linear regression model $w_2$ is given by
\begin{eqnarray*}
w_1 & = & \left(\sum_{i=1}^n M_* x_i x_i^{\top} M_* + \lambda\right)^{-1}\sum_{i=1}^n M_* x_iy_i \\
& = & \left(\sum_{i=1}^n M_* x_i x_i^{\top} M_* + \lambda\right)^{-1} M_* \left(\sum_{i=1}^n x_i x_i^{\top}\right) M_* w_* + \sum_{i=1}^n  \left(\sum_{i=1}^n M_* x_i x_i^{\top} M_* +\lambda I\right)^{-1} M_* x_i z_i \\
& = & \left(\sum_{i=1}^n P_K M_* x_i x_i^{\top} M_* P_K + \lambda\right)^{-1} P_K M_*\left(\sum_{i=1}^n  x_i x_i^{\top}\right)M_* P_K w_* + \sum_{i=1}^n  \left(\sum_{i=1}^n M_* x_i x_i^{\top} M_* +\lambda I\right)^{-1} M_* x_i z_i
\end{eqnarray*}
Using the same analysis, we have, with a probability $1 - \delta$, 
\[
\lambda_{\min}\left(\frac{1}{n}\sum_{i=1}^n P_K M_* x_i x_i^{\top} M_* P_K\right) \geq \left(1 - \sqrt{\frac{3R(\lambda_K + \gamma)}{n\lambda^2_K}\log\frac{1}{\delta}}\right)\lambda_K
\]
where $|x_i| \leq R$ for all $i \in [n]$. Hence, when
\[
n \geq \frac{12R(\lambda_K + \gamma)}{\lambda^2_K}\log\frac{1}{\delta}
\]
we have, with a probability at least $1 - \delta$, 
\[
\lambda_{\min}\left(\frac{1}{n}\sum_{i=1}^n P_K M_* x_i x_i^{\top} M_* P_K\right) \geq \frac{\lambda^2_K}{2(\lambda_K + \gamma)}
\]
and therefore
\[
|w_2 - w_*| \leq \frac{\lambda}{\lambda^2_K/2(\lambda_K + \gamma) + \lambda} w_* + \frac{1}{\lambda}\left|\frac{1}{n}\sum_{i=1}^n M_* x_i z_i\right| 
\]
Using the vector concentration inequality, we have, with a probability $1 - \delta$
\[
\left|\frac{1}{n}\sum_{i=1}^n M_* x_i z_i\right| \leq \frac{R}{n} + \sqrt{\frac{\sigma^2\mbox{tr}(M_*\Sigma M_*)}{n}\log\frac{1}{\delta}}
\]
Thus, with a probability $1 - 2\delta$, we have
\begin{eqnarray}
|w_2 - w_*| \leq \frac{\lambda}{\lambda^2_K/2(\lambda_K + \gamma) + \lambda} w_* + \frac{R}{n} + \sqrt{\frac{\gamma\mbox{tr}(M_*\Sigma M_*)}{n}\log\frac{1}{\delta}} \label{eqn:bound-2}
\end{eqnarray}

Compare the bounds in (\ref{eqn:bound-1}) and (\ref{eqn:bound-2}), we consider the case when $\lambda_1, \ldots, \lambda_d$ follow a power law, i.e. $\lambda_k = R/\sqrt{k}$. We thus have
\[
\lambda_K = \frac{R^2}{K}, \; \mbox{tr}(\Sigma) = \sum_{k=1}^d \frac{R^2}{\sqrt{k}} \leq R\sqrt{d+1}
\]
and with a probability at least $1 - 2\delta$, 
\begin{eqnarray}
|w_1 - w_*| \leq \frac{\lambda}{R^2/2\sqrt{K} + \lambda} |w_*| + \frac{R}{n} + \sqrt{\frac{\sigma^2R d^{1/2}}{n}\log\frac{1}{\delta}} \label{eqn:bound-3}
\end{eqnarray}
When $\lambda = O(1/\sqrt{n})$, we have
\[
|w_1 - w_*| \leq O\left(\frac{d^{1/4}}{n^{1/2}}\right)
\]
For the pre-training case, we have
\[
\mbox{tr}(M_*\Sigma M_*) = \sum_{k=1}^d \frac{R^6}{k^{3/2}(R^2/\sqrt{k}+ \gamma)^2} \leq R^6\sum_{k=1}^d k^{-3/2} \leq 2R^6
\]
and therefore, with a probability $1 - 2\delta$,
\[
|w_2 - w_*| \leq O\sqrt{\frac{1}{n}}
\]
indicating that pretraining helps improve the generalization error.

\end{document}